\documentclass[openacc]{rsproca_new}

\usepackage{amssymb}
\usepackage{amsmath}
\usepackage{bm}
\usepackage{subfig}
\usepackage{wrapfig}
\usepackage{lipsum}     
\usepackage{float} 
\usepackage{graphicx}           
\usepackage{url} 
\usepackage{soul}
\usepackage{enumitem}
\usepackage{lineno}
\usepackage{diagbox}
\usepackage{color}

\usepackage[ruled,vlined,linesnumbered,noresetcount]{algorithm2e}

\jname{rspa}
\Journal{Proc R Soc A\ }

\begin{document}

\title{Bayesian differential programming for robust systems identification under uncertainty}

\author{
Yibo Yang$^{1}$, Mohamed Aziz Bhouri$^{1}$ and Paris Perdikaris$^{1}$}

\address{$^{1}$ Department of Mechanical Engineering \\
  and Applied Mechanics\\
  University of Pennsylvania\\
  Philadelphia, PA 19104}

\subject{Scientific machine learning, Applied Mathematics, Bayesian statistics, Data-driven modeling}


\corres{Paris Perdikaris\\
\email{pgp@seas.upenn.edu}}

\begin{abstract}
This paper presents a machine learning framework for Bayesian systems identification from noisy, sparse and irregular observations of nonlinear dynamical systems. The proposed method takes advantage of recent developments in differentiable programming to propagate gradient information through ordinary differential equation solvers and perform Bayesian inference with respect to unknown model parameters using Hamiltonian Monte Carlo sampling. This allows an efficient inference of the posterior distributions over plausible models with quantified uncertainty, while the use of sparsity-promoting priors enables the discovery of interpretable and parsimonious representations for the underlying latent dynamics. A series of numerical studies is presented to demonstrate the effectiveness of the proposed methods including nonlinear oscillators, predator-prey systems and examples from systems biology. Taken all together, our  findings put forth a flexible and robust workflow for data-driven model discovery under uncertainty. All codes and data accompanying this manuscript are available at \url{https://bit.ly/34FOJMj}.
\end{abstract}



\begin{fmtext}
\section{Introduction}\label{sec:intro}
In the era of big data, dynamical systems discovery has received a lot of attention, primarily due to the significant growth of accessible data across different scientific disciples, including systems biology \cite{ruoff2003temperature}, bio-medical imaging \cite{kak2002principles}, fluid dynamics \cite{raissi2020hidden}, climate modeling \cite{palmer1999nonlinear}, and physical chemistry \cite{feinberg1974dynamics}. Extracting a set of features that are
\end{fmtext} 


\maketitle
 \noindent interpretable and predictive is crucial for developing physical insight and gaining a better understanding of the natural phenomena under study \cite{haller2002lagrangian}. Perhaps more importantly, this can enable the reliable forecasting of future states and subsequently lead to effective intervention strategies for design and control of complex systems \cite{tantet2018crisis, bemporad1999control, lu2019nonparametric}.

Machine learning methods and data-driven modeling techniques have already proven their utility in solving high-dimensional problems in computer vision \cite{krizhevsky2012imagenet}, natural language processing \cite{bahdanau2014neural}, etc. Due to their capability of extracting features from high dimensional and multi-fidelity noisy data  \cite{yang2019conditional, han2018solving}, these methods are also gaining attraction in modeling and simulating physical and biological systems. The evolution of such systems can be typically characterized by differential equations, and several techniques have been developed to construct predictive algorithms that can synergistically combine data and mechanistic prior knowledge. Such scientific machine learning approaches are currently employed to distill dynamics from time-series data \cite{rackauckas2020universal,gholami2019anode,chen2018neural,brunton2016discovering,rudy2017data, brennan2018data,raissi2018multistep, qin2019data}, infer the solution of differential equations \cite{raissi2017inferring, raissi2019physics, zhu2019physics, yang2019adversarial, yang2020physics,wang2020understanding}, infer parameters, latent variables and unknown constitutive laws \cite{wang2019non, deveney2019deep, raissi2020hidden, raissi2018hidden, tartakovsky2018learning}, as well as tackle forward and inverse problems in complex application domains including cardiovascular flow dynamics, \cite{kissas2020machine}, metamaterials \cite{chen2019physics}, cardiac electrophysiology \cite{sahli2020physics}, etc.

Specific to systems identification, most recent data-driven approaches \cite{rackauckas2020universal, brunton2016discovering, qin2019data, chen2018neural,gholami2019anode} heavily rely on the quality of the observations and are not designed to return predictions with quantified uncertainty. For instance, the sparse regression methods put forth in  \cite{brunton2016discovering, rudy2017data} can exhibit unstable behavior if the data is highly noisy and are not able to directly digest time-series data with irregular sampling frequency or missing values. On the other hand, recent approaches leveraging differentiable programming \cite{rackauckas2020universal,chen2018neural,gholami2019anode} can support irregularly sampled data, but are only designed to provide point-estimates for the discovered dynamics with no characterization of predictive uncertainty. Such lack of robustness and missing capabilities may limit the use of existing techniques to idealized settings and pose the need for a more flexible framework that can effectively accommodate noisy, sparse and irregularly sampled data to infer posterior distributions over plausible models, and subsequently yield robust future forecasts with quantified uncertainty.

In this work, we aim to address the aforementioned capability gaps by formulating a fully Bayesian framework for robust systems identification from imperfect time-series data. Our specific contributions can be summarized in the following points:
\begin{itemize}[leftmargin=*]
    \item We leverage recent developments in differentiable programming \cite{rackauckas2020universal,chen2018neural,gholami2019anode} to enable gradient back-propagation through numerical ODE solvers, and utilize this information to construct accelerated Hamiltonian Monte Carlo schemes for Bayesian inference.
    \item The proposed workflow is computationally efficient, end-to-end differentiable, and can directly accommodate sparse, noisy and irregularly sampled time-series data.
    \item Equipped with sparsity-promoting priors, we can recover interpretable and parsimonious representations for the latent dynamics, while entire posterior distributions over plausible models are obtained.
    \item This probabilistic formulation is key for safe-guarding against erroneous data, incomplete model parametrizations, as well as for producing reliable future forecasts with quantified uncertainty.
    \item We demonstrate enhanced capabilities and robustness against state-of-the-art methods for systems identification \cite{brunton2016discovering, rudy2017data} across a range of benchmark problems.
\end{itemize}
Taken all together, our findings put forth a novel, flexible and robust workflow for data-driven model discovery under uncertainty that can potentially lead to improved algorithms for robust forecasting, control and model-based reinforcement learning of complex systems.

The rest of this paper is organized as follows. Section \ref{sec:methods} presents the proposed method and its corresponding technical ingredients. In section \ref{sec:HMC}, a review of Bayesian inference and Hamiltonian Monte Carlo sampling is given. In section \ref{sec:Neural ODE}, we provide details on how the proposed method can effectively propagate gradient information through ODE solvers leveraging differential programming. In section \ref{sec:normalization}, we discuss a simple but essential step for pre-processing the observed data in order to obtain robust training behavior. In section \ref{sec:Results} the effectiveness of the proposed method is tested on a series of numerical studies, including nonlinear oscillators, predator-prey systems and a realistic example in systems biology. Finally, in section \ref{sec:discussion} we summarize our key findings, discuss the limitations of the proposed approach, and carve out directions for future investigation.

\section{Methods}\label{sec:methods}
This section provides a comprehensive overview of the key ingredients that define our work, namely differential programming via Neural ordinary differential equations (NeuralODEs) \cite{chen2018neural} and Bayesian inference with Hamiltonian Monte Carlo sampling \cite{neal2011mcmc}. 
Our presentation is focused on describing how these techniques are be interfaced to obtain a novel, efficient, and robust workflow for parsimonious model discovery from imperfect time-series data.

\subsection{Differentiable programming}\label{sec:Neural ODE}
In their original work, Chen {\em et. al.} \cite{chen2018neural} introduced a general framework for propagating  gradient information through classical numerical solvers for ordinary differential equations (ODEs) that blends classical adjoint methods \cite{pontryagin2018mathematical} with modern developments in automatic differentiation \cite{van2018automatic}. To illustrate the main concepts, consider a general dynamical system of the form

\begin{equation}\label{equ:dynamics}
    \frac{d \bm{x}}{dt} = f(\bm{x}, t; \bm{\theta}),
\end{equation}
where $\bm{x}\in\mathbb{R}^{D}$ denotes the state space of the $D$-dimensional dynamical system, and $\bm{\theta}$ is a vector unknown parameters that parametrizes the latent dynamics $f:\mathbb{R}^{D}\rightarrow\mathbb{R}^{D}$. A systems identification task is now summarized as follows. Given some observations $\bm{x}_{i}, i = 1, ..., n$ evaluated at time instances $t_{i}$, $i = 1, ..., n$, one would like to learn the $\bm{\theta}$ that best parametrizes the underlying dynamics. A typical approach for identifying these optimal parameters is to define a loss function $g$ that measures the discrepancy between the observed data 
$\bm{x}_{i+1}$ and the model's predictions $\hat{\bm{x}}_{i+1}$ for a given $\bm{\theta}$, i.e.,

\begin{equation}
    \mathcal{J}(\bm{\theta}) = \sum_{i=1}^{n-1}g(\bm{x}_{i+1},h_{\bm{\theta}}(\bm{x}(t_{i}))),
\end{equation}
where $h_{\bm{\theta}}(\bm{x}(t_{i}))=\hat{\bm{x}}_{i+1}$ is the predicted value under a given set of estimated model parameters $\bm{\theta}$ obtained by integrating the dynamical system with some ODE solver. $g(\cdot)$ could be any metric to evaluate the distance / evaluating the discrepancy between the true value and the predicted one (e.g., $L_1$ loss, $L_2$ loss \cite{hastie2009elements} or KL-divergence \cite{ghahramani2015probabilistic}, Wasserstein distance \cite{gulrajani2017improved}, etc). A sufficient way to minimize the loss function is through gradient descent \cite{su2014differential, bottou2010large}, however appropriate methods need to be employed for effectively computing the gradient of the loss function with respect to the parameters, namely $\frac{\partial L}{\partial\bm{\theta}}$. This is done  by defining the adjoint of the dynamical system as $\bm{a}(t) = \frac{\partial L}{\partial\bm{x}(t)}$. Then, the dynamical system describing the evolution of the adjoint can be derived as \cite{pontryagin2018mathematical, chen2018neural}

\begin{equation}
    \frac{d\bm{a}(t)}{dt} = -\bm{a}(t)^{T}\frac{\partial f(\bm{x}(t), t, \bm{\theta})}{\partial \bm{x}}.
\end{equation}

Note that the adjoint $\bm{a}(t)$ can be computed by an additional call to the chosen ODE solver, and the target derivative $\frac{\partial L}{\partial\bm{\theta}}$ can be then computed as

\begin{equation}\label{equ:adjoint}
    \frac{\partial L}{\partial\bm{\theta}} = - \int_{t_1}^{t_0}\bm{a}(t)^{T}\frac{\partial f(\bm{x}(t), t, \bm{\theta})}{\partial \bm{\theta}}dt,
\end{equation}
where $\frac{\partial f(\bm{x}(t), t, \bm{\theta})}{\partial \bm{\theta}}$ can be evaluated via automatic differentiation \cite{van2018automatic,chen2018neural}.
%
%
The main advantages of this approach can be summarized in the following points:
\begin{itemize}
  \item The data does not need to be collected on a regular time grid.
  \item The time-step $\Delta t_i$ between an observed data pair $\{\bm{x}(t_i), \bm{x}(t_i + \Delta t_i)\}$ can be relatively large. Within each $\Delta t_i$, a classical numerical scheme can be used to integrate equation \ref{equ:adjoint}, where $\Delta t_i = N_i dt$ is discretized in $N_i$ equal spaced steps, with the step size $dt$ being typically chosen according to the stability properties of the underlying ODE solver. 
  \item As this setup only assumes dependency between individual input-output pairs $\{\bm{x}(t_i), \bm{x}(t_i + \Delta t_i)\}$, the observed  time-series data does not need to be continuous and could be selected from different time intervals.
\end{itemize}
%
The specific choices of $\Delta t_i$, $dt$ and $N_i$ will be discussed for each  of the different examples given in this work. In general, the choice of $N$ is made based on the following trade-off between accuracy and computational complexity. To this end, small $N$ may lead to a less accurate model, while large $N$ would lead to a massive computational graph that can significantly slow down model training. 

The unknown model parameters $\bm{\theta}$ can be estimated by minimizing appropriate loss function. Throughout this work, the following $L_2$ loss is employed

\begin{equation}
\label{eq:loss_fn}
\begin{aligned}
    &\mathcal{J}(\bm{\theta}) = \sum_{i = 1}^{n}\|\bm{x}(t_i + \Delta t_i) - h_{\bm{\theta}}(\bm{x}(t_i))\|^2,
\end{aligned}
\end{equation}
where $h_{\bm{\theta}}(\bm{x}(t_i))$ denotes the output of a numerical ODE solver. Throughout this work, we use the classical fourth order Runge-Kutta method \cite{iserles2009first}, although more general choices can be employed \cite{rackauckas2020universal}. The training data-set consists of pairs $\{(\bm{x}(t_1), \bm{x}(t_1 + \Delta t_1)), (\bm{x}(t_2), \bm{x}(t_2 + \Delta t_2)), ..., (\bm{x}(t_n), \bm{x}(t_n + \Delta t_n))\}$. To simplify notation, let $\bm{X}(\bm{t})$ be defined as $\bm{X}(\bm{t}) = \{\bm{x}(t_1), \bm{x}(t_2), ...,\bm{x}(t_n)\}$, such that $\bm{X}(\bm{t} + \bm{\Delta t}) = \{\bm{x}(t_1 + \Delta t_1), \bm{x}(t_2 + \Delta t_2), ...,\bm{x}(t_n + \Delta t_n)\}$, then the training data-set is given by: $\mathcal{D} = \{\bm{X}(\bm{t}), \bm{X}(\bm{t} + \bm{\Delta t})\}$. 

Notice that the aforementioned workflow is only capable of producing deterministic point estimates for the unknown parameters $\bm{\theta}$, typically corresponding to a local minimum of equation \ref{eq:loss_fn}.
In many practical cases it is desirable to obtain a distribution over plausible parameter configurations that can effectively characterize the uncertainty in the estimates due to noise and sparsity in the observed data, as well as potential misspecifications in the model parametrization. A  framework for accounting for such uncertainties can be constructed by adopting a Bayesian approach via the use of effective sampling algorithms for approximating a posterior distribution over the unknown model parameters $p(\bm{\theta}|\mathcal{D})$, as discussed in the next section.


\subsection{Bayesian inference with Hamiltonian Monte Carlo}\label{sec:HMC}
The Bayesian formalism provides a natural way to account for uncertainty, while also enabling the injection of prior information for the unknown model parameters $\bm{\theta}$ (e.g., sparsity), as well as for modeling sparse and noisy observations in the training data-set. Perhaps more importantly, it enables the complete statistical characterization for all inferred parameters in the model. The latter, is encapsulated in the posterior distribution which can be factorized as
\begin{equation}
    \label{eq:unnormalized_Bayes}
     p(\gamma, \lambda, \bm{\theta}| \bm{X}(\bm{t} + \bm{\Delta t}), \bm{X}(\bm{t})) \varpropto p(\bm{X}(\bm{t} + \bm{\Delta t})|\bm{X}(\bm{t}), \bm{\theta}, \gamma) p(\bm{\theta}|\lambda) p(\gamma)p(\lambda),
\end{equation}
where $p(\bm{X}(\bm{t} + \bm{\Delta t})|\bm{X}(\bm{t}), \bm{\theta}, \gamma)$ is a likelihood function that measures the discrepancy between the observed data and the model's predictions for a given set of parameters $\bm{\theta}$, $p(\bm{\theta}|\lambda)$ is a prior distribution parametrized by a set of  parameters $\lambda$ that can help encode any domain knowledge about the unknown model parameters $\bm{\theta}$, and $\gamma$ contains a set of parameters that aim to characterize the noise process that may be corrupting the observations. In this work, a hierarchical Bayesian approach corresponding to the following likelihood and priors is employed:

\begin{equation}
\label{eq:posterior}
\begin{aligned}
    p(\bm{X}(\bm{t} + \bm{\Delta t})|\bm{X}(\bm{t}), \bm{\theta}, \gamma) & = \prod_{i = 1}^{N}\mathcal{N}(\bm{x}(t_i + \Delta t_i)|f(\bm{x}(t_i), t_i, \bm{\theta}), \gamma^{-1}), \\
    p(\bm{\theta}|\lambda) & = \textrm{Laplace}(\bm{\theta}|0, \lambda^{-1}),\\
    p(\log\lambda) & = \textrm{Gam}(\log\lambda| \alpha_1, \beta_1),\\
    p(\log\gamma) & = \textrm{Gam}(\log\gamma| \alpha_2, \beta_2).
\end{aligned}
\end{equation}
The use of a Gaussian likelihood stems from assuming a simple isotropic Gaussian noise model with zero mean and precision $\gamma$. The Laplace prior over the unknown model parameters $\bm{\theta}$ \cite{williams1995bayesian} can promote sparsity in the inferred model representations and enable an effective reduction of the influence of any irrelevant parameters -- a key feature for recovering interpretable and parsimonious representations \cite{brunton2016discovering}. The Gamma distribution is a common choice for the prior distributions over the unknown precision parameters $\lambda$ and $\gamma$. For additional motivation and alternative choices, the interested reader is referred to \cite{geweke1993bayesian, gelman2013bayesian, winkler1967assessment, bernardo1979reference, berger1990robust}. Finally, the logarithm of the precision variables $\lambda$ and $\gamma$ is used to ensure that their estimated values remain positive during model training.

The posterior distribution defined in equation \ref{eq:posterior} is not analytically tractable in general due to the modeling assumptions on the likelihood and priors, as well as due to the presence of non-linearity in the latent dynamics of equation \ref{equ:dynamics}. Typically, sampling from this unnormalized distribution is difficult and computationally expensive, especially when the dimension of $\bm{\theta}$ is large. Hamiltonian Monte Carlo (HMC) \cite{neal2011mcmc} is a powerful tool to handle Bayesian inference tasks in high dimensions by utilizing gradient information to effectively generate approximate posterior samples. To illustrate the key ideas behind HMC sampling, let us denote $\bm{\Theta} = [\bm{\theta}, \log(\lambda), \log(\gamma)]$ as the vector including all the unknown parameters that need to be inferred from data. The starting point for building an HMC sampler is to define a Hamiltonian function $H = U(\bm{\Theta}) + V(\bm{v})$, where $U(\bm{\Theta})$ is the potential energy of the original system usually taken as the logarithm of the unnormalized distribution in equation \ref{eq:posterior}, and  $V(\bm{v})$ is the kinetic energy of the system introduced by the auxiliary velocity variables $\bm{v}:=\frac{d\bm{\Theta}}{dt}$. The evolution of $\bm{\Theta}$ and $\bm{v}$ can be expressed by taking gradients of the Hamiltonian as

\begin{equation}
    \label{eq:hamiltonian_dynamics}
   \frac{d\bm{v}}{dt} = - \frac{\partial H}{\partial \bm{\Theta}}, \quad \frac{d\bm{\Theta}}{dt} = \frac{\partial H}{\partial \bm{v}}.
\end{equation}

A Markov chain that convergences to the stationary distribution of equation \ref{eq:unnormalized_Bayes} can be simulated by integrating this dynamical system using an energy preserving leapfrog scheme \cite{neal2011mcmc}. Note here that computing the gradient $\frac{\partial H(t)}{\partial \bm{\Theta}}$ is not trivial as the evaluation of the likelihood function involves the call to an ODE solver. This difficulty can be directly addressed via the differentiable programming approach discussed in section \ref{sec:Neural ODE} to effectively enable gradient back-propagation through the ODE solver. Note that all parameters in $\bm{\Theta}$ can be either updated simultaneously, or separately for $\bm{\theta}$ and $\{\lambda,\gamma\}$ using a Metropolis-within-Gibbs  scheme \cite{gilks1995adaptive, millar2000non}.



\subsection{Learning dynamics with Bayesian differential programming}
\label{sec:learning_dyanamics}
Here we distinguish between three different problem settings that cover a broad range of practical applications. The first class consists of problems in which the model form of the underlying latent dynamics is completely unknown. In this case, one can parametrize the unknown dynamical system  using black-box function approximators such as deep neural networks \cite{chen2018neural, gholami2019anode, raissi2018multistep}, or aim to distill a more parsimonious and interpretable representation by constructing a comprehensive dictionary over all possible interactions and try to infer a predictive, yet minimal model form \cite{brunton2016discovering, rudy2017data, qin2019data}. The second class of problems contains cases where a model form for the underlying dynamics is prescribed by domain knowledge, but a number of unknown parameters needs to be calibrated in order to accurately explain the observed data \cite{tartakovsky2018learning,yazdani2019systems}. Finally, a third class of problems arises as a hybrid of the aforementioned settings, in which some parts of the model form may be known from domain knowledge, but there exists additional functional terms that need to be inferred \cite{rackauckas2020universal}. As detailed in the following sections, the proposed workflow can seamlessly accommodate all of the aforementioned cases in a unified fashion, while remaining robust with respect to incomplete model parametrizations, as well as imperfections in the observed data. To this end, a general framework can be constructed by parametrizing the unknown  dynamics as
\begin{equation}\label{equ:Discovery}
\begin{aligned}
    \frac{d\bm{x}}{dt} = \underbrace{A\varphi(\bm{x})}_{\text{dictionary}} + \underbrace{f_{\bm{w}}(\bm{x})}_{\text{black-box}},\\
\end{aligned}
\end{equation}
where $A\in\mathbb{R}^{D\times K}$ represents a matrix of $D\times K$ unknown coefficients, with $K$ being the length of a dictionary $\varphi(\bm{x})\in\mathbb{R}^K$, that may or may not be constructed using domain knowledge. Specifically, $\varphi(\bm{x})$ represents the possible terms that may appear in the right hand side of the ordinary differential equation, which could encapsulate a known model form or, more generally, a prescribed dictionary of features (e.g., polynomials, Fourier modes, etc., and combinations thereof) \cite{brunton2016discovering, rudy2017data, champion2019data}. On the other hand, $f_{\bm{w}}(\bm{x})$ denotes a black-box function approximator (e.g., a neural network) parametrized by $\bm{w}$ that aims to account for any missing interactions that are not explicitly captured by $\varphi(\bm{x})$ \cite{rackauckas2020universal}.

The domain knowledge can play a crucial role in enhancing the efficiency of the resulting inference scheme as it can effectively reduce the size of the dictionary, and, consequently, the number of data required to train the model \cite{zhu2019physics, raissi2019physics}. Such knowledge is also critical to constrain the space of admissible solutions such that key physical principles are faithfully captured by the inferred  model (e.g., convergence to equilibrium limit cycles in chemical systems \cite{schnakenberg1979simple}, conservation of mass and momentum in fluid dynamics \cite{raissi2018hidden}, etc).

Although the use of dictionary learning (with or without specific domain knowledge) offers a flexible paradigm for recovering interpretable dynamic  representations, it may not always be sufficient to explain the observed data, as important terms may be missing from the model parametrization. To address this shortcoming one can try to capture these missing interactions via the use of closure terms that often lack physical intuition, and hence can be represented by a black-box function approximator $f_{\bm{w}}(\bm{x})$ with parameters $\bm{w}$, such as a deep neural network \cite{rackauckas2020universal}. Under this setup, the algorithmic framework outlined in sections \ref{sec:Neural ODE} and \ref{sec:HMC} is employed to jointly perform probabilistic  inference over plausible sets of model parameters $\bm{\theta}:=\{A,\bm{w}\}$ that yield interpretable, parsimonious, and predictive representations, as well as the precision parameters $\lambda$ and $\gamma$ of the Bayesian hierarchical model in equation \ref{eq:posterior}.

\subsection{Generating forecasts with quantified uncertainty}\label{sec:predictive_inference}
The goal of the HMC algorithm described in section \ref{sec:HMC} is to produce a faithful set of samples that concentrate in regions of high-probability in the posterior distribution $p(\bm{\theta}, \lambda, \gamma |\mathcal{D})$. Approximating this distribution is central to the proposed workflow as it enables the generation of future forecasts $\bm{x}^*(t)$ with quantified uncertainty via computing the predictive posterior distribution 

\begin{equation}\label{equ:predictive_distribution}
\begin{aligned}
    p(\bm{x}^*(t)|\mathcal{D}, \bm{x}_0, t) = \int p(\bm{x}^*(t)|\bm{\theta}, \bm{x}_0, t)p(\bm{\theta}|\mathcal{D} )d\bm{\theta}.
\end{aligned}
\end{equation}
This predictive distribution provides a complete statistical characterization for the forecasted states by encapsulating epistemic uncertainty in the inferred dynamics, as well as accounting for the fact that the model was trained on a finite set of noisy observations.
This allows the generation of plausible realizations of $\bm{x}^*(t)$ by sampling from the predictive posterior distribution as 
\begin{equation}\label{equ:predictive_approach}
\begin{aligned}
    \bm{x}^*(t) = h_{\bm{\theta}}(\bm{x}_0, t) + \epsilon, \quad  \epsilon \sim \mathcal{N}(0, \gamma^{-1}), \quad \bm{\theta},\lambda,\gamma\sim p(\bm{\theta}, \lambda, \gamma|\mathcal{D})
\end{aligned}
\end{equation}
where, $\bm{\theta}$, $\lambda$ and $\gamma$ are approximate samples from $p(\bm{\theta}, \lambda, \gamma |\mathcal{D})$ computed during model training via HMC sampling, $h_{\bm{\theta}}(\bm{x}_0, t)$ denotes any numerical integrator that takes some initial condition $\bm{x}_0$ and predicts the system's state at any time $t$, and $\epsilon$ accounts for the noise corrupting the observations used during model training. Moreover, the maximum a-posteriori (MAP) estimate of the model parameters is given as follows: 

\begin{equation}\label{equ:MAP_parameters}
\begin{aligned}
    \bm{\theta}_{\textrm{MAP}}, \lambda_{\textrm{MAP}}, \gamma_{\textrm{MAP}} = \arg\max_{\bm{\theta},\lambda,\gamma} p(\bm{\theta},\lambda,\gamma |\mathcal{D}),
\end{aligned}
\end{equation}
and it is used to obtain a point estimate prediction of the predicted states $\hat{\bm{x}}_{\textrm{MAP}}(t)$ defined as following:
\begin{equation}\label{equ:MAP_estimate}
\begin{aligned}
    \bm{x}^*_{\textrm{MAP}}(t) = h_{\bm{\theta}_{\textrm{MAP}}}(\bm{x}_0, t).
\end{aligned}
\end{equation}
Finally, it is straightforward to utilize the posterior samples of $\bm{\theta}\sim p(\bm{\theta}|\mathcal{D})$ to approximate the first- and second-order statistics of the predicted states $\bm{x}^*(t)$ for any given initial condition $\bm{x}_0$ as
\begin{align}
    \label{eq:statistics}
    \hat{\mu}_{\bm{x}^*}(\bm{x}_0, t) & = \int h_{\bm{\theta}}(\bm{x}_0, t)p(\bm{\theta}|\mathcal{D})d\bm{\theta} \approx \frac{1}{N_s}\sum\limits_{i=1}^{N_s} h_{\bm{\theta}_i}(\bm{x}_0, t),\\
    \hat{\sigma}^{2}_{\bm{x}^*}(\bm{x}_0, t) & = \int [h_{\bm{\theta}}(\bm{x}_0, t)- \hat{\mu}_{\bm{x}^*}(\bm{x}_0, t)]^2 p(\bm{\theta}|\mathcal{D})d\bm{\theta} \approx \frac{1}{N_s}\sum\limits_{i=1}^{N_s} [h_{\bm{\theta}_i}(\bm{x}_0, t) - \hat{\mu}_{\bm{x}^*}(\bm{x}_0, t)]^2,
\end{align}
where $N_s$ denotes the number of samples drawn from the Hamiltonian Markov Chain used to simulate the posterior, i.e., $\bm{\theta}_i\sim p(\bm{\theta}|\mathcal{D})$, $i=1,\dots,N_s$. Note that higher-order moments are also readily computable in a similar manner.

\subsection{Model initialization and data pre-processing}\label{sec:normalization}
To promote robustness and stability in the training of the proposed machine learning pipeline, users should be cognizant of several important aspects. 
First, although the proposed Bayesian approach can naturally safe-guard against over-fitting, it is important that a reasonable amount of training data is provided -- relative to the complexity of the system -- in order to mitigate any effects of prior misspecification. Second, the training data should be appropriately normalized in order to prevent gradient pathologies during back-propagation \cite{glorot2010understanding}. The specific utility of this step will be demonstrated in the numerical examples presented in this work, and is carried out using a standard normalization of the form
\begin{equation}
\label{eq:normalization}
    \tilde{\bm{x}} = \frac{\bm{x}}{\sigma_{\bm{x}}}
\end{equation}
where $\sigma_{\bm{x}}$ is the dimension-wise standard deviation of the training data and the division is an element-wise operation. Notice that this modification directly implies that the assumed parametrization of the underlying dynamical system also needs to be normalized accordingly (see section \ref{sec:Results} for a more detailed discussion). A third important remark here, is that the noise precision $\gamma$ obtained from the model aims to reflect the noise level in the observed data. However, it may not be the true noise level because the initial condition of the ODE can also be noisy. This point will be further discussed in section \ref{sec:Results}. 

Another important point relates to the initialization of the Hamiltonian Monte Carlo Markov Chain sampler. To this end, in order to mitigate poor mixing and convergence to local minima, a preconditioning step is considered to find a reasonable initial guess for the unknown variables $\bm{\theta}$ that parametrize the underlying latent dynamics. This step is typically carried out by minimizing the reconstruction loss of the training data using an $L_1$ regularization, 
\begin{equation}
\label{eq:preconditioning}
\mathcal{L}(\bm{\theta}) = \frac{1}{n}\sum\limits_{i=1}^{n}||\bm{x}_{i} - \hat{\bm{x}_{i}}\|^2 + \beta\|\bm{\theta}\|,
\end{equation}
using a small number ($\mathcal{O}(10^3)$) of stochastic gradient descent iterations. Notice that this essentially aims at obtaining a rough point estimate for $\bm{\theta}$, where the use of $L_1$ regularization stems from employing a sparsity-promoting Laplace prior. This preconditioning step is closely related to the SINDy algorithm of Brunton {\em et. al.} \cite{brunton2016discovering}, however without the limitations of requiring training data with small noise amplitude and sampled on regular time grid with small a time step. Moreover, the numerical experiments carried out indicate that tuning the hyper-parameter $\beta$ has almost no effect on the obtained results for all the problems considered in this work, as this simply serves as an initialization step for the HMC sampler. Hence, in all examples considered, $\beta$ is taken equal to $1$. The minimization of equation \ref{eq:preconditioning} is carried out using stochastic Adam updates \cite{kingma2014adam} with a learning rate of $\mathcal{O}(10^{-2})$. 

Note that this preconditioning is not precisely equivalent to the MAP estimation of the posterior distribution over all model parameters $\bm{\Theta}$, since the initialization of the precision parameters $\lambda$ and $\gamma$ follows a different treatment. Specifically, 
for all the problems considered in section \ref{sec:Results}, the parameters $\alpha_i$'s and $\beta_i$'s of the prior Gamma distributions are chosen to be $1$. Moreover, the precision of the Gaussian noise distribution $\gamma$ is initialized as follows. If the preconditioning precision is larger than $\exp(6)$, which means the training data appears to be nearly noise-free, the initial guess for $\gamma$ is set to $\exp(6)$ to avoid numerical stagnancy of the HMC sampler. Otherwise, if the preconditioning precision is less than $\exp(6)$, then it is used as the initial guess for $\gamma$. This empirical initialization strategy has a positive effect in accelerating the convergence of the HMC sampler for all the examples considered in section \ref{sec:Results}. 
In all of the numerical examples, the Hamiltonian Monte Carlo step-size is taken as $\epsilon = 10^{-4}$, while the number number of leapfrog steps to integrate the Hamiltonian dynamics in equation \ref{eq:hamiltonian_dynamics} is fixed to $L =10$. While this choice of $\epsilon$ is the safest choice, one can increase its value as long as not too many samples are rejected during model training. Alternatively, more sophisticated HMC samplers that allow for adaptively tuning the step-size can be employed \cite{hoffman2014no}.
Finally, in all examples the Markov Chains are simulated for $5,000$ steps, while the last $N_s= 2,000$ samples produced by HMC are used to compute the response statistics (see equation \ref{eq:statistics}).



\section{Results}
\label{sec:Results}
In this section, a comprehensive collection of numerical studies that aim to illustrate the key contributions of this work is presented and placed in context of the existing SINDy framework of Brunton {\em et. al.} \cite{brunton2016discovering}, which is currently considered as a state-of-the-art method for dictionary learning of dynamical systems. Specifically, we expand on four benchmark problems that cover all possible cases discussed in section \ref{sec:learning_dyanamics} in terms of parametrizing the latent dynamics using a dictionary, domain knowledge, black-box approximations, or a combination thereof. The algorithmic settings used across all cases follow the discussion provided in \ref{sec:normalization}, unless otherwise noticed. All code and data presented in this section will be made publicly available at \url{https://github.com/PredictiveIntelligenceLab/BayesianDifferentiableProgramming}.


\subsection{Dictionary learning for a two-dimensional nonlinear  oscillator}\label{sec:Damped}
Let us start with a pedagogical example on dictionary learning for inferring the dynamics of a two-dimensional damped oscillator from scattered time-series data \cite{raissi2018multistep}. The exact system dynamics are given by
\begin{equation}\label{eq:spiral}
\begin{aligned}
    \frac{d x_1}{dt} = \alpha x_1^3 + \beta x_2^3, \\
    \frac{d x_2}{dt} = \gamma x_1^3 + \delta x_2^3.
\end{aligned}
\end{equation}
The goal here is to recover this dynamical system directly from data using a dictionary parametrization containing polynomial terms with up to 3rd order interactions, taking the form

\begin{equation}
\begin{bmatrix}
    \frac{dx_1}{dt}\\
    \frac{dx_2}{dt}
\end{bmatrix}
= A\varphi(\bm{x}) = 
\begin{bmatrix}
  a_{11} & a_{12} & a_{13} & a_{14} & a_{15} & a_{16} & \textcolor{red}{a_{17}} & a_{18} & a_{19} & \textcolor{red}{a_{110}} \\
  a_{21} & a_{22} & a_{23} & a_{24} & a_{25} & a_{26} & \textcolor{red}{a_{27}} & a_{28} & a_{29} & \textcolor{red}{a_{210}}
\end{bmatrix}\varphi(\bm{x})
\end{equation}
where $\varphi(\bm{x}) = [1, x_1, x_2, x_1^2, x_1 x_2, x_2^2, x_1^3, x_1^2 x_2, x_1 x_2^2, x_2^3]^T$ and the $a_{ij}$'s are unknown scalar coefficients that will be estimated using the proposed Bayesian differential programming method. The model's active non-zero parameters are highlighted with red color for clarity. The goal is to infer a posterior distribution for the parameters $\bm{\theta}:=\{A\}$, while it is obvious that the coefficient matrix $A$ and dictionary $\varphi(\bm{x})$ will increase in size with the order of the polynomial features used to parametrize the system. In all presented experiments, a set of training data is generated by simulating the exact dynamics of equation \ref{eq:spiral} in the time interval $t\in [0, 20]$ with the initial condition $(x_1, x_2) = (2, 0)$ and with $\alpha = -0.1, \beta = 2.0, \gamma = -2.0$, and $\delta = -0.1$. In what follows, the performance of the proposed algorithms is investigated with respect to the temporal resolution and the noise level in the observed data. Moreover, the analysis is provided with a comprehensive comparison with the SINDy algorithm of Brunton {\em et. al.} \cite{brunton2016discovering}.

\subsubsection{Effects of sparsity in the training data}

To examine the performance of the proposed methods with respect to sparsity in the training data, a data-set is generated by integrating the exact dynamics with a relatively large $dt = 0.0677$, such that there are only $n=300$ training data pairs. In order to establish a comparison against the SINDy algorithm \cite{brunton2016discovering}, the training data is assumed noise-free and sampled on a regular temporal grid, as required by the SINDy setup \cite{brunton2016discovering}.

\begin{figure}
\centering
\includegraphics[width=\textwidth]{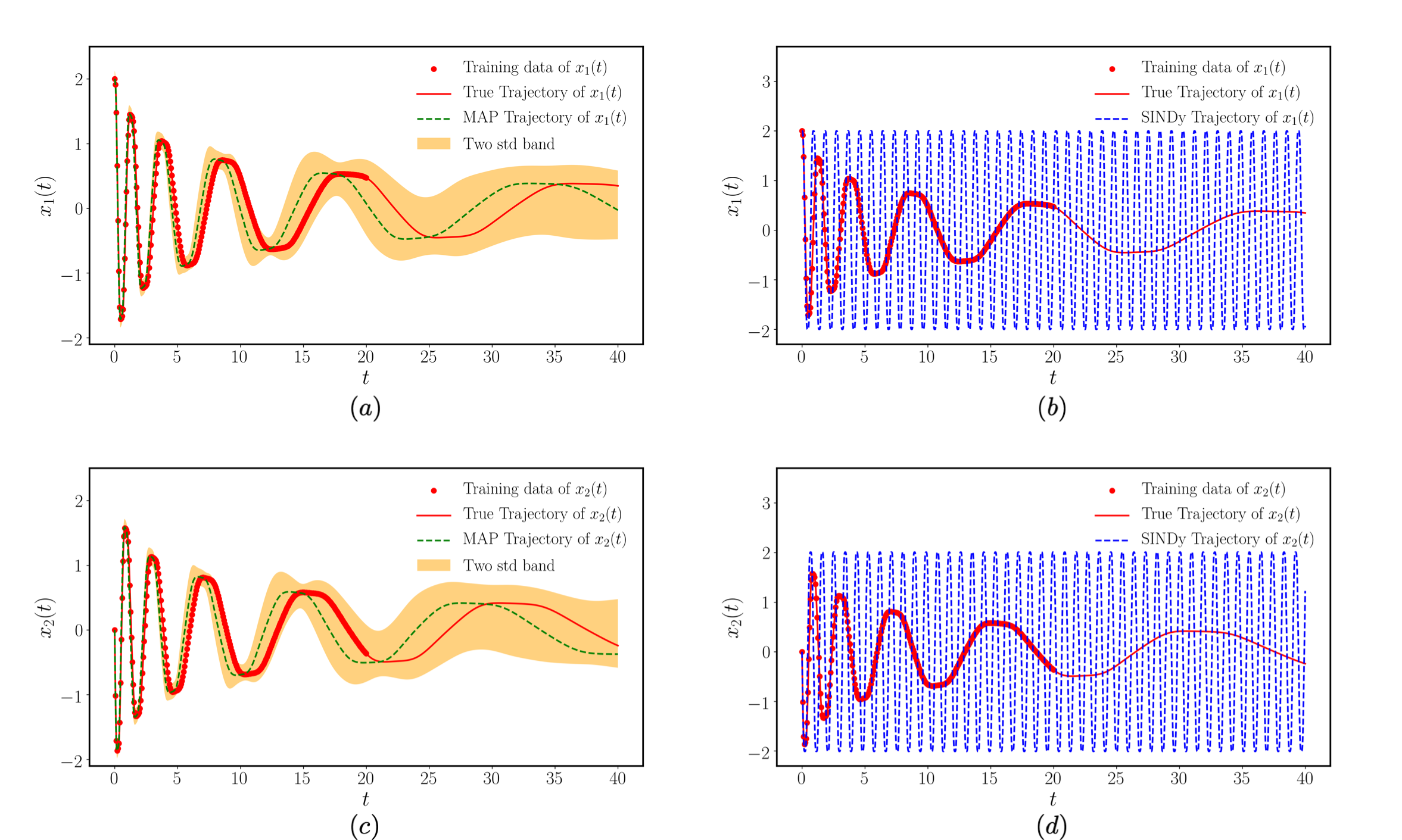}
\caption{{\em Two-dimensional damped oscillator with low-resolution training data:} (a) Learned dynamics versus the true dynamics and the training data for $x_1(t)$. (b) SINDy's prediction for $x_1(t)$ versus the true dynamics and the training data. (c) Learned dynamics versus the true dynamics and the training data for $x_2(t)$. (d) SINDy's prediction for $x_2(t)$ versus the true dynamics and the training data.}
\label{fig:Spiral_Large_dt_compare}
\end{figure}

\begin{figure}
\centering
\includegraphics[width=\textwidth]{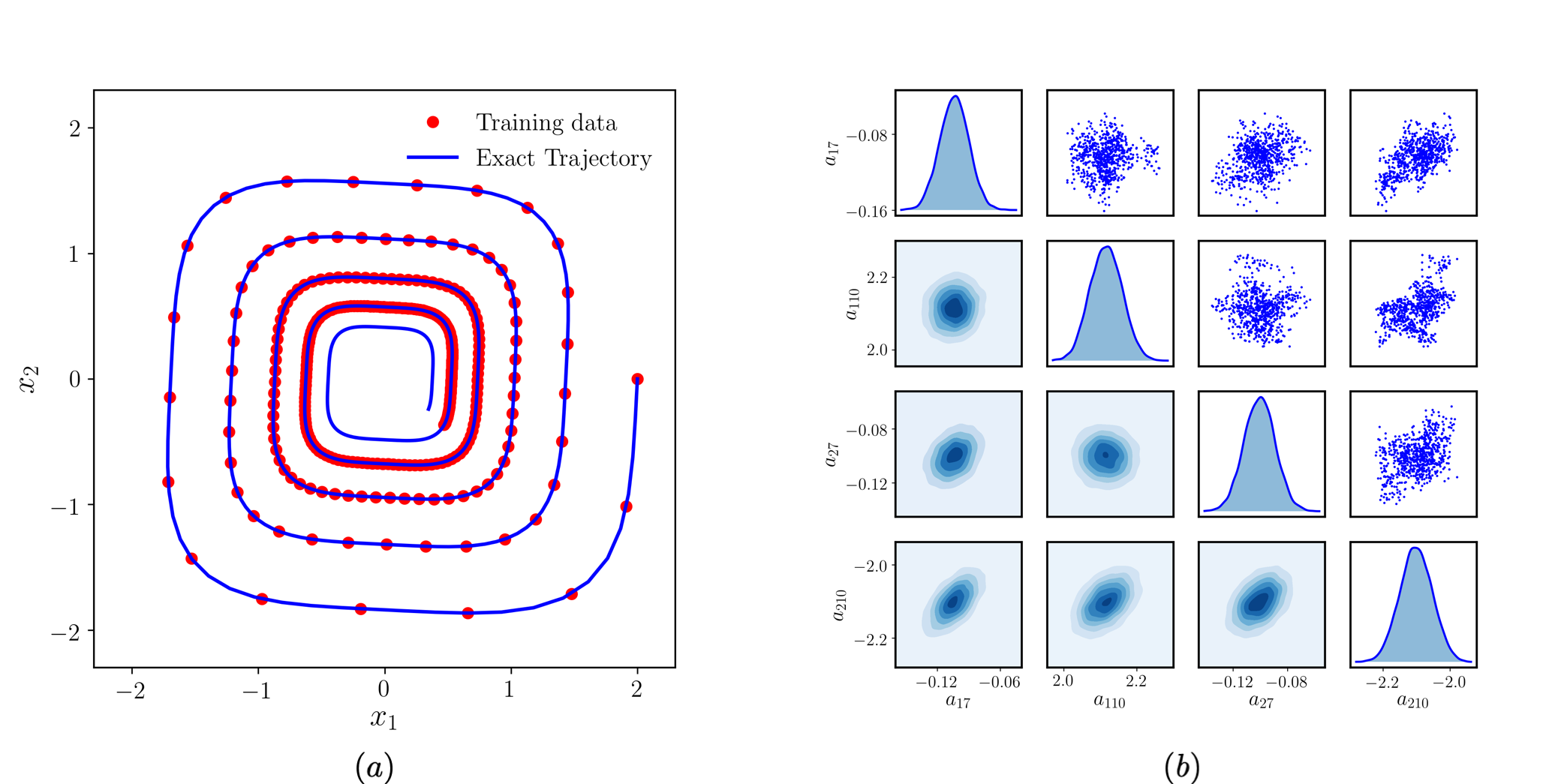}
\caption{{\em Two-dimensional damped oscillator with low-resolution training data:} (a) Phase plot of the training data and the true trajectory. (b) Posterior distribution of the inferred active model parameters.}
\label{fig:Spiral_Large_dt_data}
\end{figure}

As seen in figures \ref{fig:Spiral_Large_dt_compare} and \ref{fig:Spiral_Large_dt_data}, the SINDy algorithm fails to identify the underlying system, while the proposed approach remains robust thanks to the Bayesian formulation outlined in section \ref{sec:HMC} and which enables a faithful statistical characterization of the latent system dynamics even under sparse observations. Indeed, the proposed method does not only have the capability of accurately identifying the parameters even for relatively large $dt$ (see table \ref{tab:Spiral_large_dt}), but also provides reasonable uncertainty estimates for the extrapolated long-term forecasts. In contrast, the SINDy algorithm gives inaccurate estimations for the dictionary parameters (see table \ref{tab:Spiral_large_dt_SINDy}), consequently leading to large errors in the forecasted system states. As shown in figures \ref{fig:Spiral_Large_dt_compare}(b),(d), the estimated trajectories clearly deviate from the exact solution since its oscillatory frequency is considerably higher than the exact one, and it is not capturing the decay of the oscillations' peak. Moreover, SINDy's results highly depend on the choice of the sequential least-squares threshold parameter value, and slightly different values of this hyper-parameter may give significantly different results. 

\begin{table}[!h]
\centering
\caption{{\em Dictionary learning for a two-dimensional nonlinear  oscillator:} MAP estimation of the inferred model parameters using low-resolution training data ($dt = 0.0677$).}
\label{tab:Spiral_large_dt}
\begin{tabular}{lllllllllllllll}
\hline
$a_{11}$ & $a_{12}$ & $a_{13}$ & $a_{14}$ & $a_{15}$ & $a_{16}$ & $\textcolor{red}{a_{17}}$ & $a_{18}$ & $a_{19}$ & $\textcolor{red}{a_{110}}$ \\
\hline
-0.0052 & 0.0057 & -0.0095 & 0.0094 & 0.0116 & 0.0057 & $\textcolor{red}{-0.097}$ & -0.0093 & -0.054 & $\textcolor{red}{2.092}$\\
\hline 
$a_{21}$ & $a_{22}$ & $a_{23}$ & $a_{24}$ & $a_{25}$ & $a_{26}$ & $\textcolor{red}{a_{27}}$ & $a_{28}$ & $a_{29}$ & $\textcolor{red}{a_{210}}$\\
\hline
0.0048& 0.0195 & -0.0072 & -0.021 & 0.0024 & -0.0075 & $\textcolor{red}{-2.08}$ & 0.0280 & -0.014 & $\textcolor{red}{-0.099}$\\
\hline
\end{tabular}
\vspace*{-4pt}
\end{table}


\begin{table}[!h]
\centering
\caption{{\em Dictionary learning for a two-dimensional nonlinear  oscillator:} Point estimates for the dictionary coefficients obtained by the  SINDy algorithm \cite{brunton2016discovering} using low-resolution training data ($dt = 0.0677$).}
\label{tab:Spiral_large_dt_SINDy}
\begin{tabular}{lllllllllllllll}
\hline
$a_{11}$ & $a_{12}$ & $a_{13}$ & $a_{14}$ & $a_{15}$ & $a_{16}$ & $\textcolor{red}{a_{17}}$ & $a_{18}$ & $a_{19}$ & $\textcolor{red}{a_{110}}$ \\
\hline
  0& 0 & 0 & 0 & 0 & 0 & \textcolor{red}{0} & 0 & 0 & \textcolor{red}{2.10}\\
\hline 
$a_{21}$ & $a_{22}$ & $a_{23}$ & $a_{24}$ & $a_{25}$ & $a_{26}$ & $\textcolor{red}{a_{27}}$ & $a_{28}$ & $a_{29}$ & $\textcolor{red}{a_{210}}$\\
\hline
  0& 0 & 0 & 0 & 0 & 0 & \textcolor{red}{-1.87} & 0 & 0 & \textcolor{red}{0} \\
\hline
\end{tabular}
\vspace*{-4pt}
\end{table}

\subsubsection{Effects of noise in the training data}

\begin{figure}
\centering
\includegraphics[width=\textwidth]{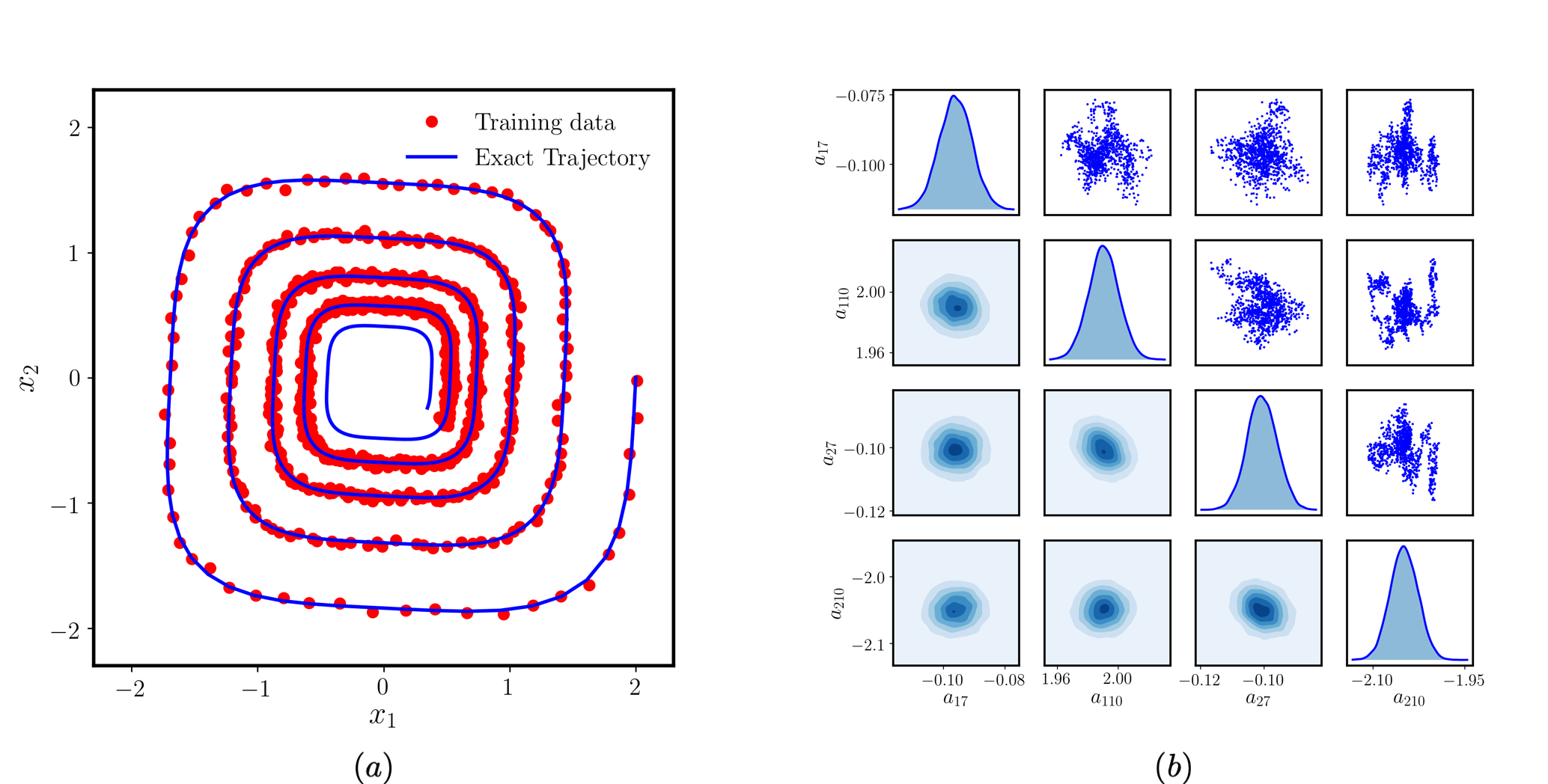}
\caption{{\em Two-dimensional damped oscillator with noisy training data:} (a) Phase plot of the training data and the true trajectory. (b) Posterior distribution of the inferred active model parameters.}
\label{fig:Spiral_noisy_data}
\end{figure}

In this section, the sensitivity of the proposed methods with respect to the presence of noise in the training data is investigated. To this end, a training data-set is generated with $n=1,000$ equi-spaced data-pairs in $t\in [0, 20]$ by simulating the exact dynamics of equation \ref{eq:spiral} with $dt = 0.02$, and the observations are deliberately corrupted with uncorrelated Gaussian noise of the form $\mathcal{N}(0, 0.02^2)$ (see figure \ref{fig:Spiral_noisy_data}(a)). 

\begin{figure}
\centering
\includegraphics[width=\textwidth]{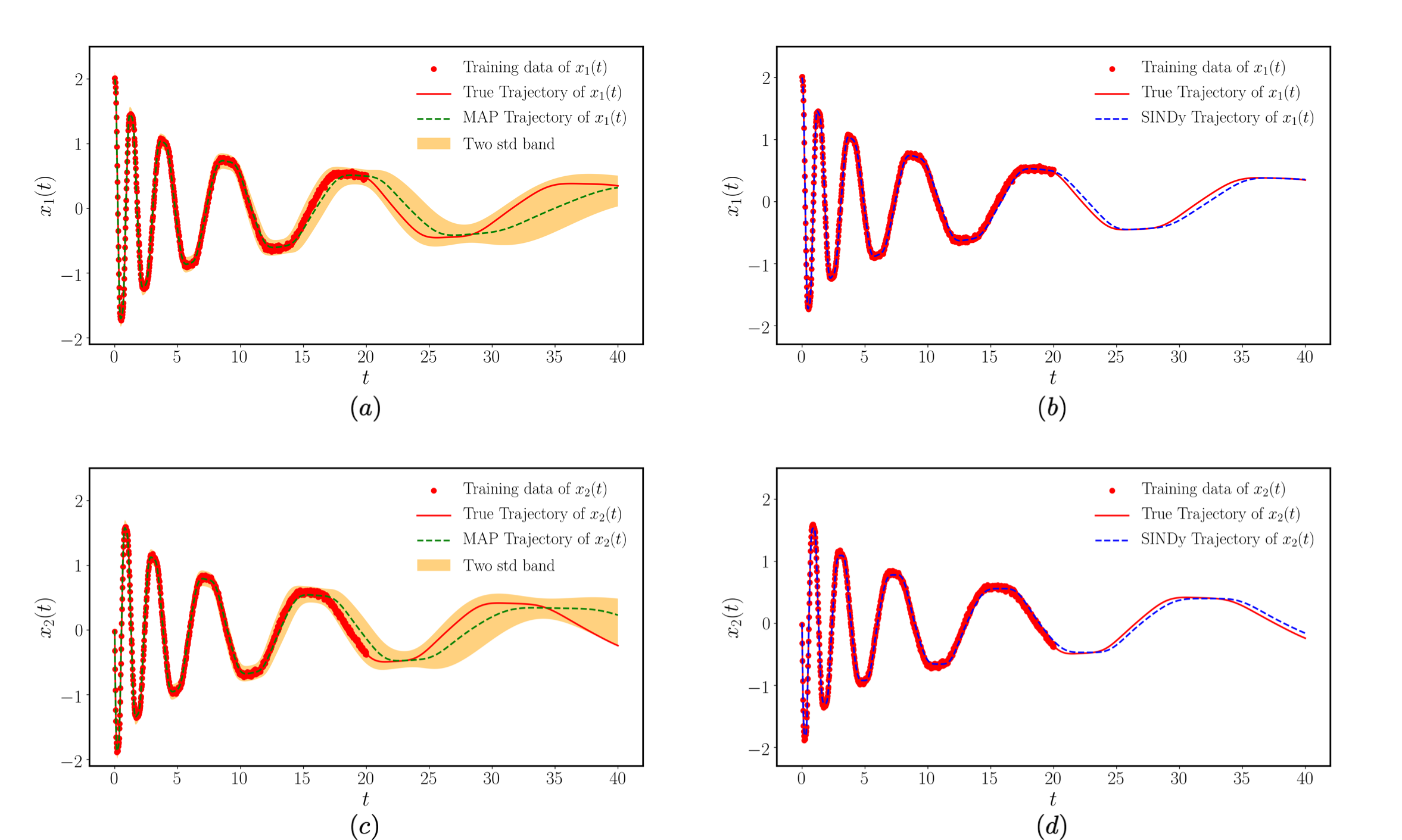}
\caption{{\em Two-dimensional damped oscillator with noisy training data:} (a) Learned dynamics versus the true dynamics and the training data for $x_1(t)$. (b) SINDy's prediction for $x_1(t)$ versus the true dynamics and the training data. (c) Learned dynamics versus the true dynamics and the training data for $x_2(t)$. (d) SINDy's prediction for $x_2(t)$ versus the true dynamics and the training data.}
\label{fig:Spiral_noisy_compare}
\end{figure}

Figure \ref{fig:Spiral_noisy_compare} summarizes the predictions of the proposed Bayesian framework in comparison to the SINDy algorithm of Brunton {\em et. al.} \cite{brunton2016discovering}. Moreover, the inferred dictionary parameters are provided for both methods in tables \ref{tab:Spiral_noisy} and \ref{tab:Spiral_noisy_SINDy}, respectively. Notice that, although the predicted trajectories of SINDy are quite close to the true trajectories, the identified parameters are quite different from the true model form. Even though SINDy employs a total variation diminishing (TVD) regularization to safe-guard against small noise corruptions in the training data \cite{brunton2016discovering}, it is still prone to providing inaccurate results in cases where the training data is imperfect. This limitation is addressed in the proposed framework by explicitly accounting for the effects of noise in the training data using the hierarchical Bayesian model in equation \ref{eq:unnormalized_Bayes}. This source of uncertainty is also effectively propagated through the system's dynamics to yield a sensible characterization of uncertainty in the predicted forecasts (see figure \ref{fig:Spiral_noisy_compare}(a)), via the inferred posterior distribution over the model parameters (see figure \ref{fig:Spiral_noisy_data}(b)). Moreover, the resulting MAP estimates for the model parameters exhibit excellent agreement with the ground truth, as reported in table \ref{tab:Spiral_noisy}.
This result illustrates the robust performance of the proposed framework in identifying interpretable and parsimonious system representations, even in the presence of noise in the training data. 


\begin{table}[!h]
\centering
\caption{{\em Dictionary learning for a two-dimensional nonlinear oscillator:} MAP estimation of the inferred model parameters using noisy training data.}
\label{tab:Spiral_noisy}
\begin{tabular}{lllllllllllllll}
\hline
$a_{11}$ & $a_{12}$ & $a_{13}$ & $a_{14}$ & $a_{15}$ & $a_{16}$ & $\textcolor{red}{a_{17}}$ & $a_{18}$ & $a_{19}$ & $\textcolor{red}{a_{110}}$ \\
\hline
  0.0042& -0.0059 & -0.0045 & -0.0072 & -0.0078 & -0.0071 & \textcolor{red}{-0.10} & 0.047 & 0.014 & \textcolor{red}{2.0}\\
\hline 
$a_{21}$ & $a_{22}$ & $a_{23}$ & $a_{24}$ & $a_{25}$ & $a_{26}$ & $\textcolor{red}{a_{27}}$ & $a_{28}$ & $a_{29}$ & $\textcolor{red}{a_{210}}$\\
\hline
  -0.0029& -0.010 & -0.0073 & 0.0017 & 0.017 & 0.0064 & \textcolor{red}{-2.0} & 0.015 & -0.049 & $\textcolor{red}{-0.108}$ \\
\hline
\end{tabular}
\vspace*{-4pt}
\end{table}

\begin{table}[!h]
\centering
\caption{{\em Dictionary learning for a two-dimensional nonlinear oscillator:} estimation of the SINDy's parameters using noisy training data.}
\label{tab:Spiral_noisy_SINDy}
\begin{tabular}{lllllllllllllll}
\hline
$a_{11}$ & $a_{12}$ & $a_{13}$ & $a_{14}$ & $a_{15}$ & $a_{16}$ & $\textcolor{red}{a_{17}}$ & $a_{18}$ & $a_{19}$ & $\textcolor{red}{a_{110}}$ \\
\hline
  0& 0 & 0 & 0 & 0 & 0 & \textcolor{red}{-0.139} & 0 & 0 & \textcolor{red}{2.0}\\
\hline 
$a_{21}$ & $a_{22}$ & $a_{23}$ & $a_{24}$ & $a_{25}$ & $a_{26}$ & $\textcolor{red}{a_{27}}$ & $a_{28}$ & $a_{29}$ & $\textcolor{red}{a_{210}}$\\
\hline
  0& 0 & 0 & 0 & 0 & 0 & \textcolor{red}{-1.97} & 0 & -0.188 & \textcolor{red}{0} \\
\hline
\end{tabular}
\vspace*{-4pt}
\end{table}


\subsection{Parameter inference in a predator-prey system with irregularly sampled observations}\label{sec:LV}
This case study is designed to illustrate the capability of the proposed framework to accommodate noisy and irregularly sampled time-series data; a common practical setting that cannot be effectively addressed by SINDy and other popular data-driven systems identification methods \cite{brunton2016discovering,rudy2017data,lusch2018deep,raissi2018multistep, qin2019data}. To this end, a classical prey-predator system described by the Lotka–Volterra equations is considered
\begin{equation}\label{eq:LV}
\begin{aligned}
    \frac{d x_1}{dt} = \alpha x_1 - \beta x_1 x_2 \ ,\\
    \frac{d x_2}{dt} = \delta x_1 x_2 - \gamma x_2 \ ,
\end{aligned}
\end{equation}
which is known to exhibit a stable limit cycle behavior for $\alpha = 1.0, \beta = -0.1, \gamma = -1.5$, and $\delta = 0.75$. Without loss of generality, a dictionary that precisely contains the active terms of the system is considered. However, the $n=1000$ training data-pairs will be irregularly sampled in the interval $t\in [0, 25]$ by randomly sub-sampling an exact trajectory of system \ref{eq:LV} starting from an initial condition set to $(x_1, x_2) = (5, 5)$.
Moreover, the training data is perturbed by $3\%$ white noise proportional to its standard deviation. Given this irregular and noisy training data, the goal is to demonstrate the performance of the proposed Bayesian framework in identifying the unknown model parameters $\bm{\theta}:=\{\alpha,\beta,\gamma,\delta\}$ with quantified uncertainty, as well as in producing sensible forecasts of extrapolated future states.

\begin{figure}
\centering
\includegraphics[width=\textwidth]{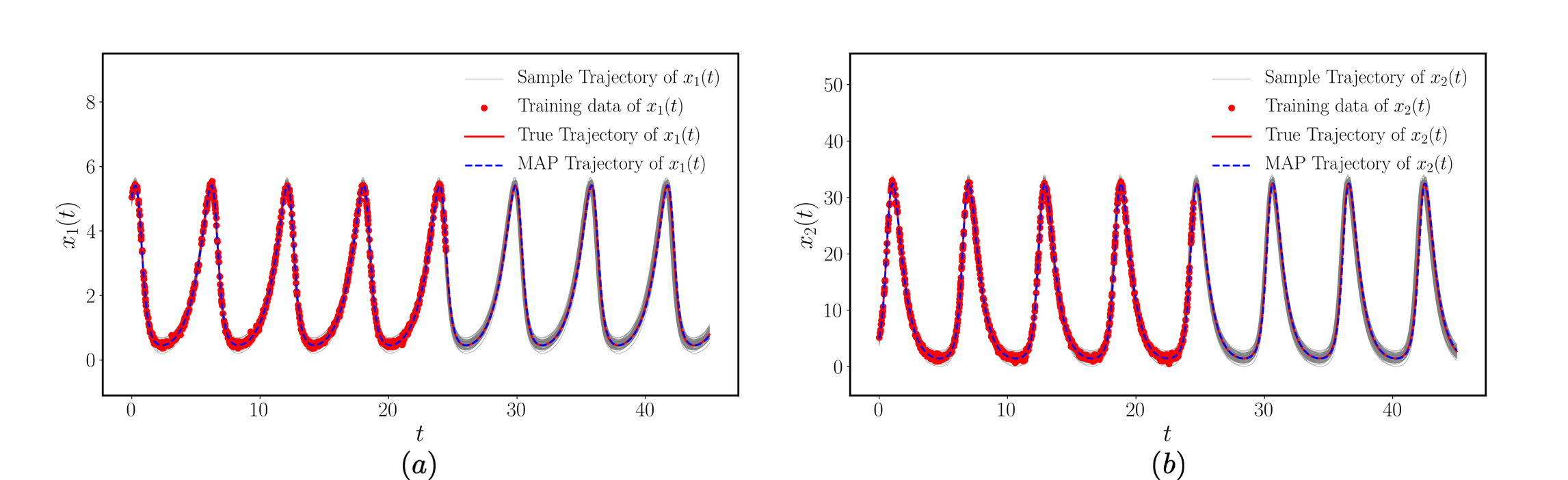}
\caption{{\em Parameter inference in a predator-prey system from irregularly sampled, noisy data:} (a) Learned dynamics versus the true dynamics and the training data for $x_1$. (b) Learned dynamics versus the true dynamics and the training data for $x_2$.}
\label{fig:LV_noisy_compare}
\end{figure}
\begin{figure}
\centering
\includegraphics[width=\textwidth]{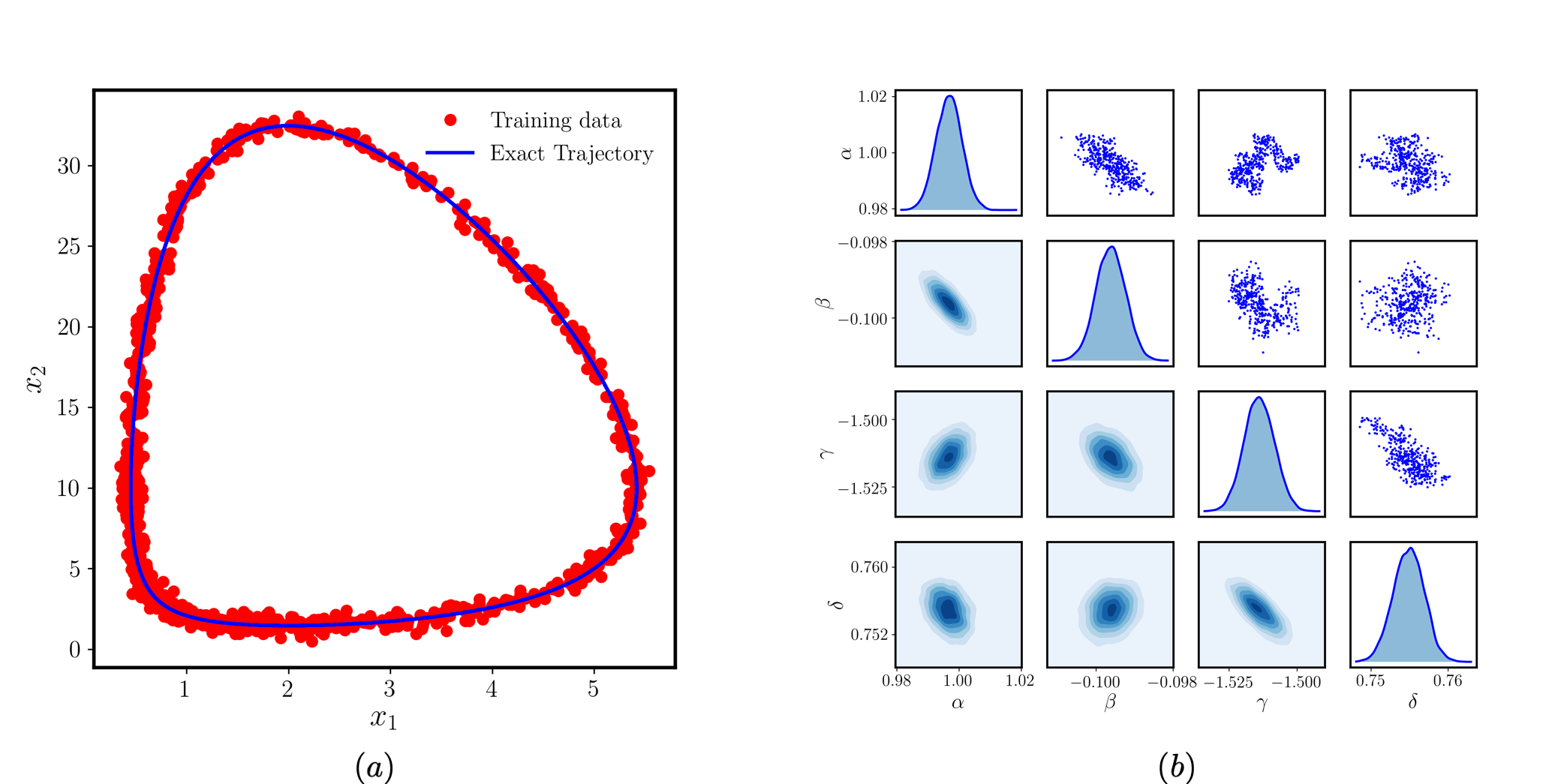}
\caption{{\em Parameter inference in a predator-prey system from irregularly sampled, noisy data:} (a) Phase plot of the training data and the true trajectory. (b) Posterior distribution of the inferred model parameters.}
\label{fig:LV_noisy_data}
\end{figure}

The results of this experiment are summarized in figures \ref{fig:LV_noisy_compare} and \ref{fig:LV_noisy_data}. It is evident that the proposed Bayesian differential programming approach (i) is able to provide an accurate estimation for the unknown model parameters, (ii) yields a MAP estimator with a predicted trajectory that closely matches the exact system's dynamics, (iii) returns a posterior distribution over plausible models that captures both the epistemic uncertainty of the assumed parametrization and the uncertainty induced by training on a finite amount of noisy training data, and (iv) propagates this uncertainty through the system's dynamics to characterize variability in the predicted future states.

Another interesting observation here is that the Hamiltonian Monte Carlo sampler is very efficient in identifying the importance/sensitivity of each inferred parameter in the model. For instance, less important parameters have the highest uncertainty, as observed in the posterior density plots shown in figures 
\ref{fig:LV_noisy_data}(b). Specifically, notice how the posterior distribution of $\alpha$ and $\gamma$ has a considerably larger standard deviation than the other parameters, implying that the evolution of this dynamical system is less sensitive with respect to these parameters.



\subsection{Safe-guarding against model inadequacy: a damped pendulum case study}\label{sec:Damped_pendulum}
The purpose of this example is to demonstrate the effects of a misspecified model parametrization, and to highlight how the proposed Bayesian framework can help detect such problematic cases and safe-guard against them. Such cases may arise when domain knowledge is insufficient to guide the selection of a parsimonious model form, as well as when important interaction terms may be missing from a dictionary representation. To illustrate the main ideas, a simple damped pendulum system described by the following equation is considered: 
\begin{equation}
\label{eq:Damped_pendulum}
\begin{aligned}
    \frac{d x_1}{dt} &= \gamma x_2,\\
    \frac{d x_2}{dt} &= - \alpha x_2 - \beta \sin(x_1),
\end{aligned}
\end{equation}
with $\gamma = 1$, $\alpha = 0.2$ and $\beta = 8.91$. A set of sparse and irregularly sampled training data-pairs can be generated by  randomly sub-sampling a simulated trajectory of the exact dynamics in $t\in[0, 20]$, starting from an initial condition $(x_1, x_2) = (-1.193, -3.876)$. This imperfect data-set can be the used to recover an interpretable model representation via dictionary learning, albeit here we  deliberately choose to use an incomplete dictionary containing polynomial terms only up to 1st order, i.e.,
\begin{equation}
\begin{bmatrix}
    \frac{dx_1}{dt}\\
    \frac{dx_2}{dt}
\end{bmatrix}
= A\varphi(\bm{x}) = 
\begin{bmatrix}
  a_{11} & a_{12} & \textcolor{red}{a_{13}} \\
  a_{21} & \textcolor{blue}{a_{22}} & \textcolor{red}{a_{23}} 
\end{bmatrix}
\begin{bmatrix}
  1 \\ x_1 \\ x_2  
\end{bmatrix},
\end{equation}
where the unknown model parameters are $\bm{\theta}:=\{a_{11}, a_{12}, a_{13}, a_{21}, a_{22}, a_{23}\}$, with the active terms being marked with red color for clarity. Moreover, a blue marker is used to highlight a mismatched term in the incomplete dictionary, hinting its erroneous capacity to approximate the true $\sin(x_1)$ term using just $x_1$ as a feature. It is evident that such a dictionary choice cannot faithfully capture the exact physics of the problem, and is hence destined to yield inaccurate predictions.
Nevertheless, here we argue that this discrepancy between the true form and the assumed incomplete parametrization of the dynamical system can be effectively detected by the proposed Bayesian workflow via inspecting the inferred posterior distribution over all parameters $p(\bm{\theta}, \lambda, \gamma|\mathcal{D})$, which is expected to exhibit high entropy in presence of a misspecified model parametrization and imperfect training data.

\begin{figure}
\centering
\includegraphics[width=\textwidth]{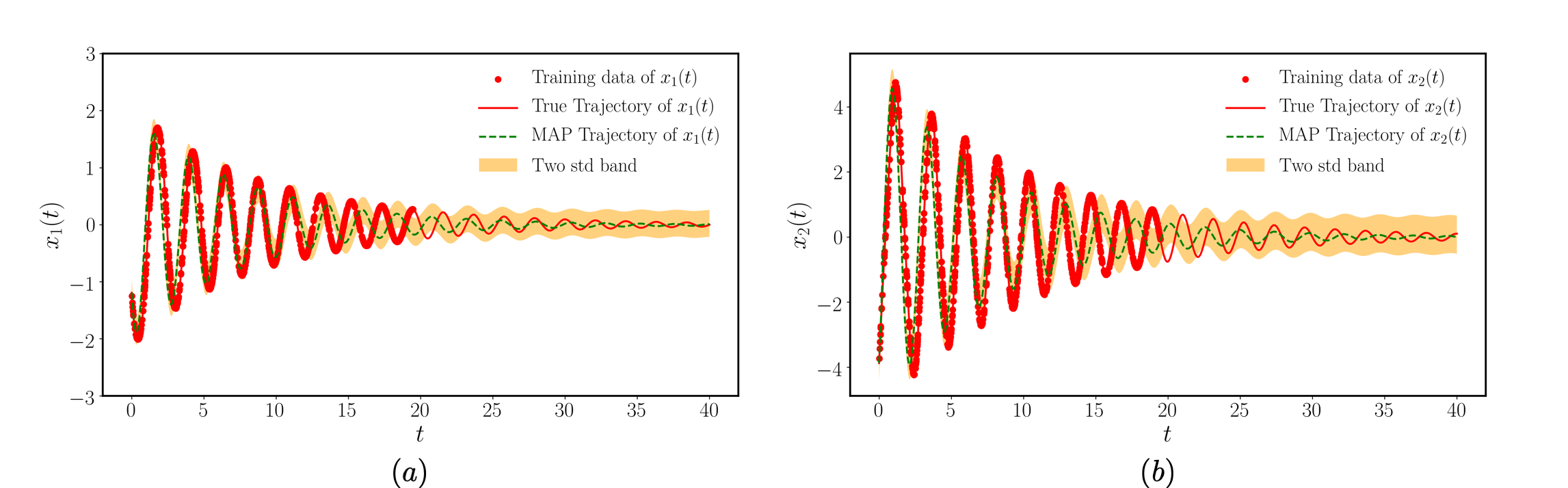}
\caption{{\em Damped pendulum with irregularly sampled data:} (a) Learned dynamics versus the true dynamics and the training data for $x_1(t)$. (b) Learned dynamics versus the true dynamics and the training data for $x_2(t)$.}
\label{fig:Damped_pendulum_compare}
\end{figure}

\begin{figure}
\centering
\includegraphics[width=\textwidth]{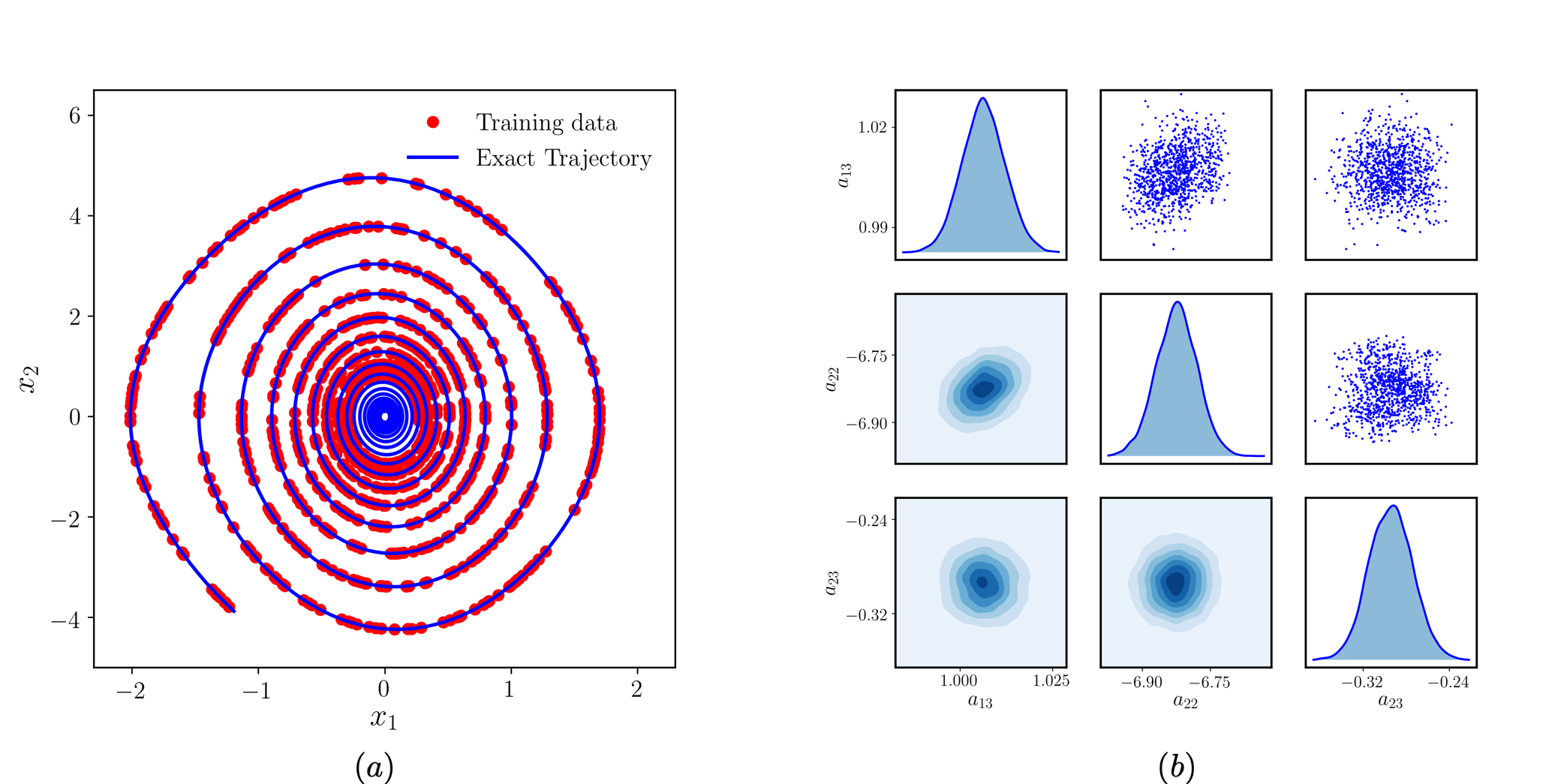}
\caption{{\em Damped pendulum with irregularly sampled data:} (a) Phase plot of the training data and the true trajectory. (b) Posterior distribution of the model's parameters.}
\label{fig:Damped_pendulum_data}
\end{figure}

The results of this experiment are summarized in figures  \ref{fig:Damped_pendulum_compare} and \ref{fig:Damped_pendulum_data}. In particular, figure \ref{fig:Damped_pendulum_compare} shows the scattered training data, the exact simulated trajectory, the predicted trajectory corresponding to the MAP estimate of the inferred model parameters, as well as a two standard deviations band around a set of plausible forecasted states. As expected, the use of a misspecified dictionary leads to an inferred model that has difficulty in fitting the observed data and yields predictions with high variability. This model inadequacy is also clearly captured in the posterior distribution over the inferred parameters shown in figure \ref{fig:Damped_pendulum_data}. Indeed, the inferred density for the $\beta$ parameter exhibits very high variance, as the $\beta$ coefficient crucially corresponds to the misspecified term in the dictionary. This is a direct indication that the inferred model is inadequate to capture the observed reality, leading to forecasts with inevitably large error-bars. This innate ability of the proposed framework to detect model misspecification via a rigorous quantification of predictive uncertainty can prove crucial in risk-sensitive applications, and provides an important capability that is currently missing from the recently active literature on data-driven model discovery  \cite{brunton2016discovering,rudy2017data, champion2019data, qin2019data, rackauckas2020universal, raissi2018multistep}.

A natural question to now ask is: can we still recover an accurate predictive model even if the true dynamics can only be partially captured by the assumed dictionary representation? To tackle this question we turn our attention to the hybrid learning setting discussed in section \ref{sec:learning_dyanamics}, in which some parts of the model can be captured by sparsely selecting interpretable terms from a dictionary, while other missing parts or closure terms can be accounted for via a black-box function approximator. To this end, the damped pendulum case study is revisited, and the dictionary learning parametrization is endowed with the ability to approximate the missing $\sin(x_1)$ term via a deep neural network as

\begin{equation}
\begin{bmatrix}
    \frac{dx_1}{dt}\\
    \frac{dx_2}{dt}
\end{bmatrix}
= A\varphi(\bm{x}) + f_{\bm{w}}(\bm{x})\\
=
\begin{bmatrix}
  a_{11} + a_{12}x_1 + \textcolor{red}{a_{13}}x_2 + a_{14} x_1^2 + a_{15}x_1 x_2 + a_{16}x_2^2  \\
  \textcolor{red}{a_{21}}x_2 + a_{22} x_1 x_2 +  a_{23}x_2^2 + f_{\bm{w}}(x_1),
\end{bmatrix}
\end{equation}
where $A$ is a matrix of unknown coefficients corresponding to a dictionary $\varphi(\bm{x})$ containing polynomial interactions up to 2nd order, and $f_{\bm{w}}(x_1)$ is a fully-connected deep neural network with $2$ hidden layers of dimension $20$, a hyperbolic tangent activation, and a set of unknown weight and bias parameters denoted by $\bm{w}$.
Notice that the neural network admits only $x_1$ as an input to make sure that the resulting parametrized dynamical system has a unique solution.
Under this setup, the proposed Bayesian framework can be employed to jointly infer the dictionary and the neural network parameters that define the model representation, namely $\bm{\theta}:=\{a_{11}, a_{12}, a_{13}, a_{14}, a_{15}, a_{16}, a_{21}, a_{22}, a_{23},\bm{w}\}$. This defines a 490-dimensional inference problem.

\begin{figure}
\centering
\includegraphics[width=\textwidth]{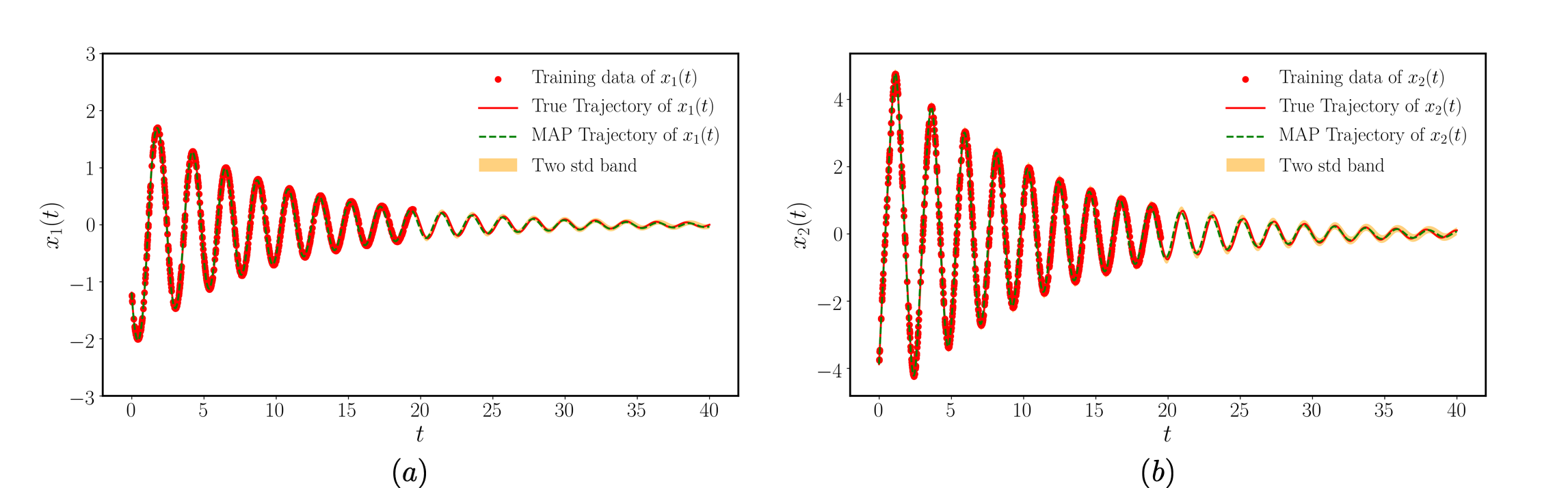}
\caption{{\em Safe-guarding against model inadequacy in damped pendulum representation:} (a) Learned dynamics versus the true dynamics and the training data for $x_1(t)$. (b) Learned dynamics versus the true dynamics and the training data for $x_2(t)$. A deep neural network is used to approximate any interactions missing from an incomplete dictionary representation.}
\label{fig:Damped_pendulum_compare_hybrid}
\end{figure}

\begin{figure}
\centering
\includegraphics[width=\textwidth]{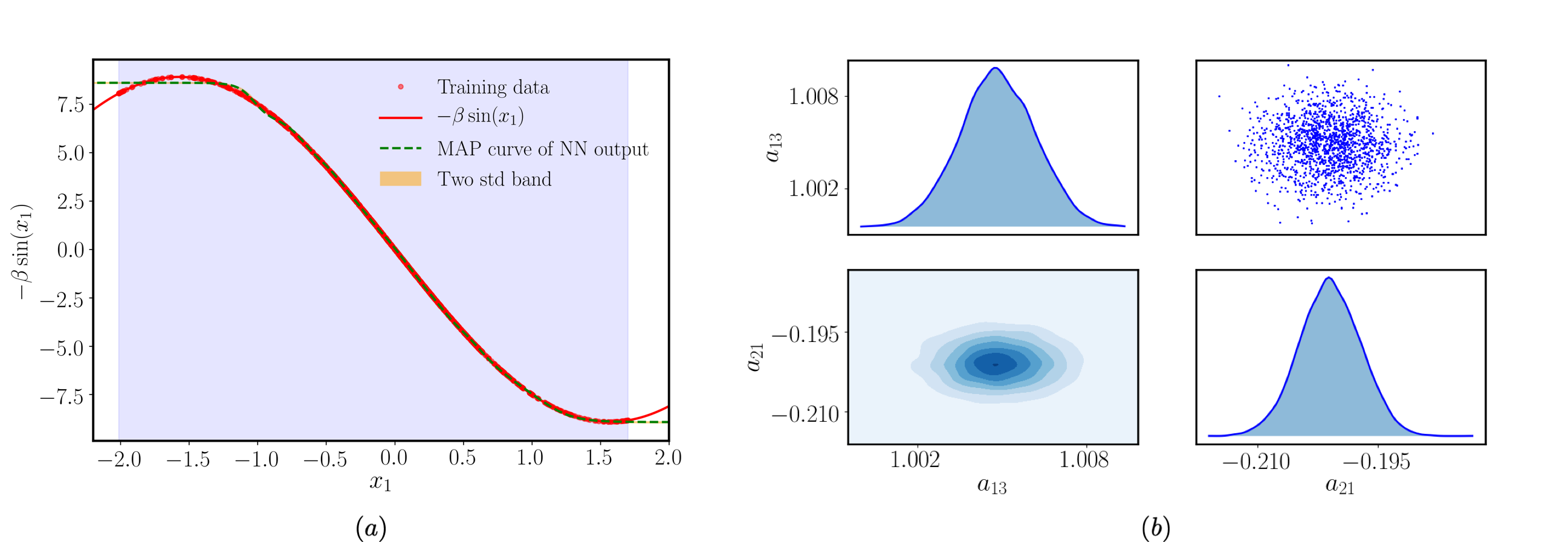}
\caption{{\em Safe-guarding against model inadequacy in damped pendulum representation:} (a) Deep neural network approximation to the missing dictionary term $-\beta\sin(x_1)$. The shade region indicated the range of data seen during model training. (b) Posterior distribution of the inferred active model parameters.}
\label{fig:Hybrid_learning_param}
\end{figure}

Figure \ref{fig:Damped_pendulum_compare_hybrid} shows the scattered training data, the exact simulated trajectory, the predicted trajectory corresponding to the MAP estimate of the inferred model parameters, as well as a two standard deviations band around a set of plausible forecasted states. It is evident that the revised hybrid model formulation can now correctly identify the true underlying dynamics, leading to an accurate predicted MAP trajectory, while the predicted uncertainty effectively diminishes and concentrates around the ground truth. Moreover, the inferred coefficients of the irrelevant terms are very close to zero, while the active terms are all  properly identified. This is also true for the neural network approximation of the missing closure term $-\beta \sin(x_1)$, as depicted in figure \ref{fig:Hybrid_learning_param} which also includes uncertainty estimates over the neural network outputs.
This simple example illustrates the great flexibility of the proposed Bayesian framework in seamlessly distilling parsimonious and interpretable models from imperfect data via physics-informed dictionary learning, as well harnessing the power of black-box representations for approximating missing closure terms. 

\subsection{Bayesian calibration of a Yeast Glycolysis model}\label{sec:Glycolysis}
In this final example, the performance of the proposed algorithms applied to a realistic problem in systems biology is investigated. To this end, a yeast glycolysis process is considered and described by a 7-dimensional dynamical system \cite{ruoff2003temperature, yazdani2019systems} as:
\begin{equation}\label{equ:Glycolysis_system}
\begin{aligned}
    \frac{d S_1}{dt} &= J_0 - k_1 S_1 A_3 [1 + (\frac{A_3}{K_I})^q]^{-1}, \\ 
    \frac{d S_2}{dt} &= 2k_1 S_1 A_3 [1 + (\frac{A_3}{K_I})^q]^{-1} - k_2 S_2 N_1 - k_6 S_2 N_2, \\
    \frac{d S_3}{dt} &= k_2 S_2 N_1 - k_3 S_3 A_2, \\ 
    \frac{d S_4}{dt} &= k_3 S_3 A_2 - k_4 S_4 N_2, - \kappa(S_4 - S_{4}^{ex}), \\
    \frac{d N_2}{dt} &= k_2 S_2 N_1 - k_4 S_4 N_2, - k_6 S_2 N_2, \\ 
    \frac{d A_3}{dt} &= - 2k_1 S_1 A_3 [1 + (\frac{A_3}{K_I})^q]^{-1} + 2k_3 S_3 A_2 - k_5 A_3, \\
    \frac{d S_{4}^{ex}}{dt} &= \phi \kappa(S_4 - S_{4}^{ex}) - k S_{4}^{ex},
\end{aligned}
\end{equation}
where $N_1 + N_2 = N$ and $A_2 + A_3 = A$. 
%
%
This model form is assumed to be known from existing domain knowledge, and the goal is (i) to compute the posterior distribution over the unknown model parameters $\bm{\theta}:=\{J_0,  k_1,  k_2, k_3, k_4, k_5, k_6, k, \kappa, q, K_I, \phi, N, A\}$ from a small set of noisy and irregularly sampled observations, and (ii) to test the ability of the inferred model to accurately generalize from different initial conditions that were not observed during model training.

A training data-set is generated by randomly sub-sampling $n=1,000$ irregularly distributed observations from a single simulated trajectory of the system from an initial condition: $(S_1, S_2, S_3, S_4, N_2, A_3, S_{4}^{ex}) = (0.5, 1.9, 0.18, 0.15, 0.16, 0.1, 0.064)$ in the time interval $t\in[0, 5]$, assuming a ground truth set of parameters obtained from the experimental data provided in \cite{ruoff2003temperature}: $J_0 = 2.5$mM/min, $k_1 = 100.0$mM/min, $k_2 = 6.0$mM/min, $k_3 = 16.0$mM/min, $k_4 = 100.0$mM/min, $k_5 = 1.28$/min, $k_6 = 12.0$mM/min, $k = 1.8$/min, $\kappa = 13.0$/min, $q = 4.0$, $K_I = 0.52$mM, $N = 1.0$mM, $A = 4.0$mM and $\phi = 0.1$. Moreover, the training data is perturbed by a $2\%$ white noise proportional to its standard deviation. 

\begin{figure}
\centering
\includegraphics[width=\textwidth]{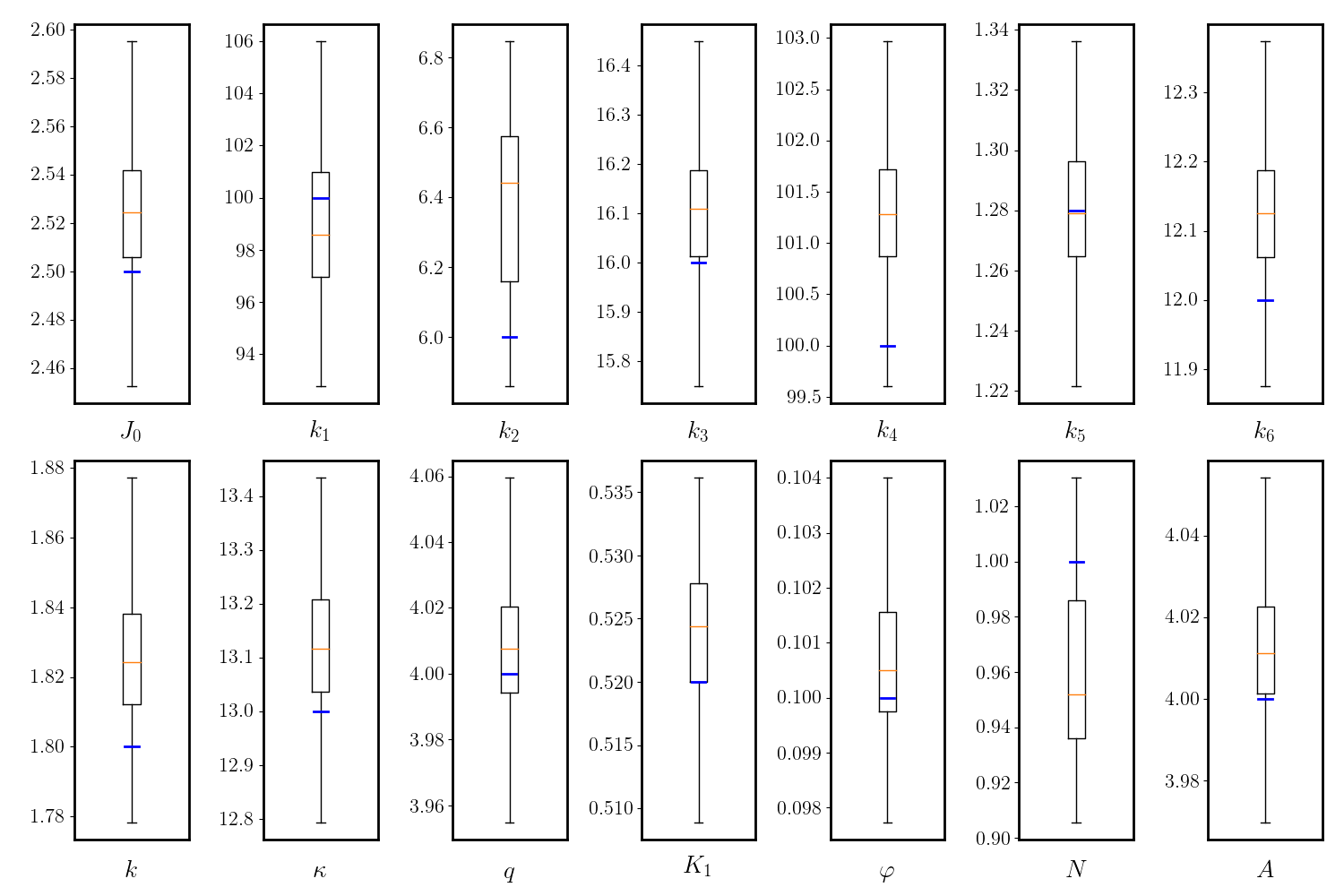}
\caption{{\em Yeast Glycolysis dynamics:} Uncertainty estimation of the inferred model parameters obtained using the proposed Bayesian differential programming method. Estimates for the minimum, maximum, median, first quantile and third quantile are provided, while the true parameter values are highlighted in blue.}
\label{fig:box_plot_Yeast_Glycolysis}
\end{figure}

Table \ref{tab:Glycolysis_params} summarizes the inferred MAP estimators for the unknown model parameters. Based on those results, all inferred parameters closely agree with the ground truth values used to generate the training data as reported in \cite{ruoff2003temperature}. Uncertainty estimates for the inferred parameters can also be deducted from the computed posterior distribution $p(\bm{\theta}|\mathcal{D})$, as presented in the box plots of figure \ref{fig:box_plot_Yeast_Glycolysis} where the minimum, maximum, median, first quantile and third quantile obtained from the HMC simulations for each parameter are presented. Finally, notice that all true values fall between the predicted quantiles, while the MAP estimators of all the parameters have considerably small relative errors compared with the true parameter values, as shown in table \ref{tab:Glycolysis_params}. 



\begin{table}[!h]
\caption{{\em Yeast Glycolysis dynamics:} MAP estimation of the inferred model parameters using noisy training data.}
\label{tab:Glycolysis_params}
\begin{tabular}{lllllllllllllll}
\hline
$J_0$ & $k_1$ & $k_2$ & $k_3$ & $k_4$ & $k_5$ & $k_6$ & $k$ & $\kappa$ & $q$ & $K_I$ & $\phi$ & $N$ & $A$ \\ 
\hline
2.52 & 98.92 & 6.43 & 16.12 & 101.24 & 1.28 & 12.13 & 1.82 & 13.12 & 4.01 & 0.52 & 0.01 & 0.95 & 4.01\\
\hline
\end{tabular}
\vspace*{-4pt}
\end{table}

Finally, to illustrate the generalization capability of the inferred model with respect to different initial conditions than those used during  training, the quality of the predicted states is assessed considering a random initial condition of $[0.428, 1.42, 0.11, 0.296, 0.252, 0.830, 0.064]$, that has not been observed during model training. 
The close agreement with the reference solution indicates that the inferred model is well capable of generalizing both in terms of handling different initial conditions, as well as extrapolating to reliable future forecasts with quantified uncertainty.

\begin{figure}
\centering
\includegraphics[width=\textwidth]{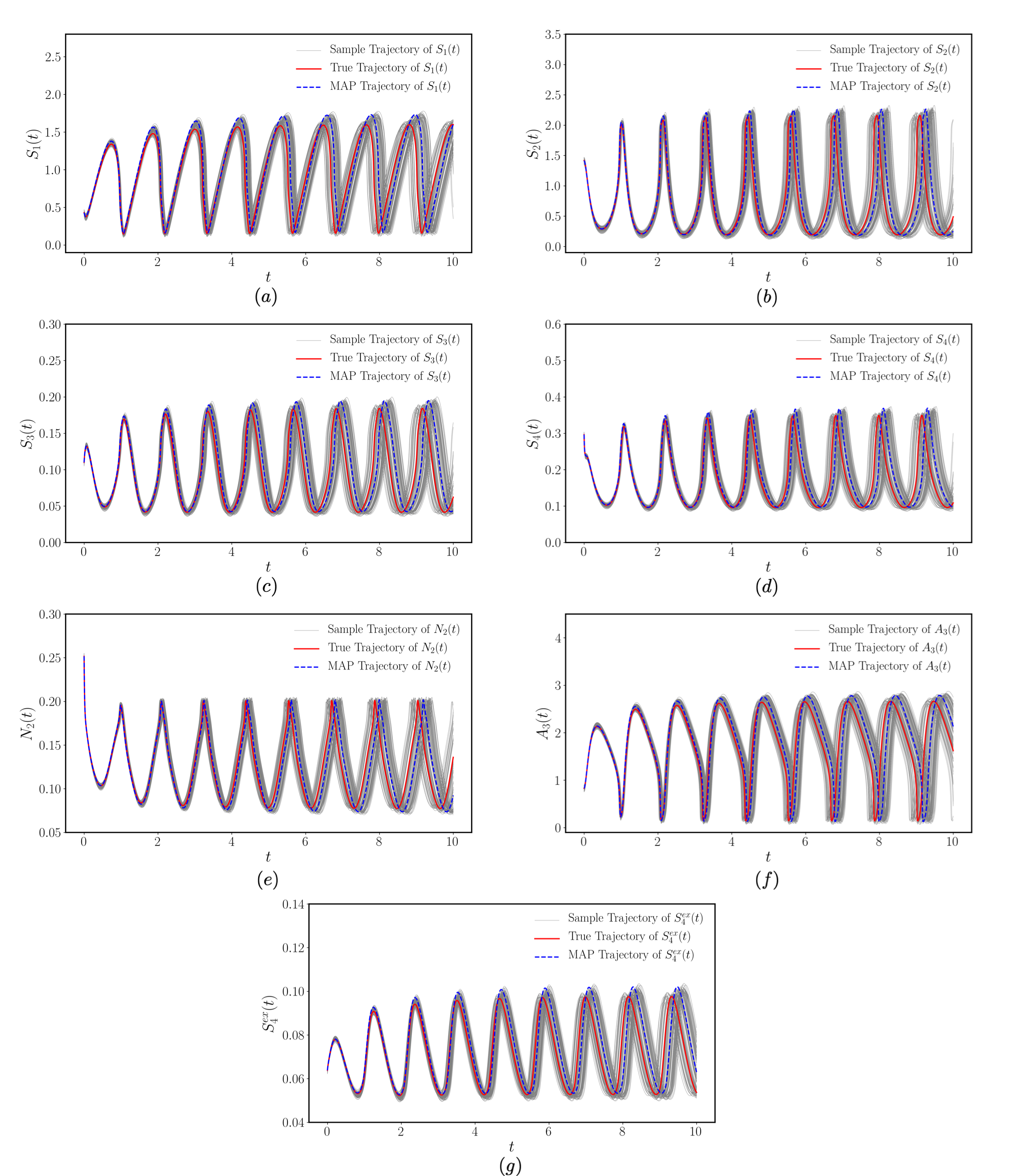}
\caption{{\em Learning Yeast Glycolysis dynamics from noisy data:} Future forecasts with quantified uncertainty from a previously unseen initial condition (i.e. an initial condition that was not used during model training).}
\label{fig:Glycolysis_noisy_different_initial}
\end{figure}

\section{Conclusions}\label{sec:discussion}
We have presented a novel machine learning framework for robust systems identification under uncertainty. The proposed framework leverages state-of-the-art differential programming techniques in combination with  gradient-enhanced sampling schemes for scalable Bayesian inference of high-dimensional posterior distributions, to infer interpretable and parsimonious representations of complex dynamical systems from {\em imperfect} (e.g. sparse, noisy, irregularly sampled) data. The developed methods are general as they can seamlessly combine dictionary learning, domain knowledge and black-box approximations, all in a computationally efficient workflow with end-to-end uncertainty quantification. The effectiveness of the proposed techniques has been systematically investigated and compared to state-of-the-art approaches across a collection of prototype problems. 


Although the proposed Bayesian differential programming framework provides great flexibility to infer a distribution over plausible parsimonious representations of a dynamical system, a number of technical issues need to be further investigated. The first relates to devising more effective initialization procedures for Markov Chain Monte Carlo sampling. Here we have partially addressed this via the MAP preconditioning algorithm discussed in section \ref{sec:normalization}, however during the preconditioning process, since the form of the dynamical system is unknown, the intermediate estimations of the parameters may cause the system to become stiff. Moreover, for cases where only very sparse observations are available, the model needs to be integrated with a large time-step $dt$ and stiffness of the system can lead to numerical instabilities during model training. A possible enhancement in this direction is to use more general stiffly stable ODE solvers as discussed in \cite{rackauckas2020universal, gholami2019anode} or more sophisticated time-step annealing strategies \cite{wang2020understanding}.  
Another potential future work could be identifying the uncertainty in model's parameters with partial observations, such that some variables of the system are not accessible. Such task would involve physics-informed regularization on the unknown latent dynamics of the system. Approaches used in \cite{yazdani2019systems} could be helpful for solving this problem. A third open question is how to adapt the proposed method to stochastic dynamical systems where the dynamics itself may be driving by a stochastic process. Approaches mentioned in \cite{li2020scalable} could be useful. Finally, the proposed Bayesian differential programming framework can be extended to parameter identification for partial differential equations (PDEs). The latter generally translates into a high dimensional dynamical systems after discretization. The learning task in this context could be carried out not only for the PDEs' parameters, but also for the discretization scheme \cite{rackauckas2020universal}.



\dataccess{All code and data accompanying this manuscript is available at  \url{https://github.com/PredictiveIntelligenceLab/BayesianDifferentiableProgramming}}.

\aucontribute{P.P. and Y.Y. conceived the methods, Y.Y. implemented the methods, Y.Y. and M.A.B. performed the simulations. Y.Y., M.A.B. and P.P. wrote the manuscript.}

\competing{The authors have no competing interests to declare.}

\funding{This work received support from the US Department of Energy under the Advanced Scientific Computing Research program (grant DE-SC0019116) and the Defense Advanced Research Projects Agency under the Physics of Artificial Intelligence program (grant HR00111890034).}




\bibliographystyle{RS} 

\bibliography{RSPA_Author_tex} 

\end{document}